\documentclass[10pt]{wlscirep}
\usepackage[utf8]{inputenc}
\usepackage[T1]{fontenc}
\usepackage{multirow,multicol,booktabs}
\usepackage{subcaption}
\usepackage{amsmath,txfonts,amssymb,amsfonts}
\usepackage[mathcal]{eucal}
\usepackage{array}
\usepackage{pdfpages}
\usepackage[square,sort,comma,numbers]{natbib}

\title{Democratising Clinical AI through Dataset Condensation for Classical Clinical Models}

\author[1,*]{Anshul Thakur}
\author[1]{Soheila Molaei}
\author[1]{Pafue Christy Nganjimi}
\author[1]{Joshua Fieggen}
\author[1]{Fenglin Liu}
\author[1,2]{Andrew A. S. Soltan}
\author[3]{Danielle Belgrave}
\author[1,4]{Lei Clifton}
\author[1,5]{David A. Clifton}
\affil[1]{University of Oxford, Oxford, UK.}
\affil[2]{Oxford University Hospitals NHS Foundation Trust, Oxford, UK.}
\affil[3]{GlaxoSmithKline, London, UK}
\affil[4]{Nuffield Department of Primary Care Health
Sciences, University of Oxford, Oxford, UK.}
\affil[5]{Oxford-Suzhou Centre for Advanced Research, Suzhou, China.}
\affil[*]{corresponding author. E-mail: \tt anshul.thakur@eng.ox.ac.uk}

\begin{abstract}
Dataset condensation (DC) learns a compact synthetic dataset that enables models to match the performance of full-data training, prioritising utility over distributional fidelity. While typically explored for computational efficiency, DC also holds promise for healthcare data democratisation, especially when paired with differential privacy, allowing synthetic data to serve as a safe alternative to real records. However, existing DC methods rely on \textit{differentiable} neural networks, limiting their compatibility with widely used clinical models such as decision trees and Cox regression. We address this gap using a differentially private, zero-order optimisation framework that extends DC to \textit{non-differentiable} models using only function evaluations. Empirical results across six datasets, including both classification and survival tasks, show that the proposed method produces condensed datasets that preserve model utility while providing effective differential privacy guarantees---enabling model-agnostic data sharing for clinical prediction tasks without exposing sensitive patient information.
\end{abstract}

\begin{document}

\flushbottom
\maketitle

\section{Introduction}
High-quality clinical data fuels progress in machine-learning-driven medicine, with electronic health records (EHRs), disease registries, and biobank datasets serving as essential resources for developing machine learning (ML) and artificial intelligence (AI) systems that support diagnosis, prognosis, and treatment planning \cite{rajkomar2019machine}. Despite their central role in model development, access to these datasets remains tightly constrained by data protection regulations and institutional governance, reflecting the sensitive nature of patient information \cite{price2019privacy,thakur2024data}. These constraints have two key consequences. First, they slow algorithmic innovation by limiting researchers’ ability to develop and evaluate models using real-world data. This hinders the development of generalisable and clinically robust systems \cite{topol2019high}. Second, they reinforce global inequities in clinical ML, with particularly acute effects in settings such as low- and middle-income countries (LMICs), where legal, infrastructural, and financial barriers to data access are often more pronounced \cite{ciecierski2022artificial}. This imbalance reinforces disparities in who benefits from clinical ML and digital health innovation, and ultimately hampers progress toward global healthcare democratisation \cite{ibrahim2021health}.

\vspace{0.1cm}
\noindent Recent studies have positioned dataset condensation (DC) as a promising paradigm to address these access constraints and enable data democratisation without compromising patient privacy \cite{wang2023medical,dong2022privacy,kanagavelu2024medsynth}. DC involves synthesising a compact set of artificial samples from the real dataset---often just a fraction of its original size---such that models trained on the synthetic data achieve performance comparable to those trained on the full dataset \cite{zhao2023dataset,guo2024towards}, while dramatically reducing storage demands and accelerating downstream training. Critically, each condensed sample is a learned aggregation of multiple real patient records, breaking direct ties to individual data points in the original dataset and providing inherent, albeit informal, privacy protection \cite{dong2022privacy,wang2023medical}. These protections can be further strengthened with differential privacy, a technique that adds carefully calibrated noise to data or computations to provide formal guarantees that individual-level information cannot be inferred \cite{carlini2022no}. In addition to enhancing privacy, DC enables more efficient and scalable model development: its minimal footprint supports faster experimentation, lowers computational barriers to training, and allows institutions without access to large-scale infrastructure to participate in clinical ML development.

\begin{figure}[h]
\centering
\includegraphics[scale=0.365, trim=1.25cm 0 0 0, clip]{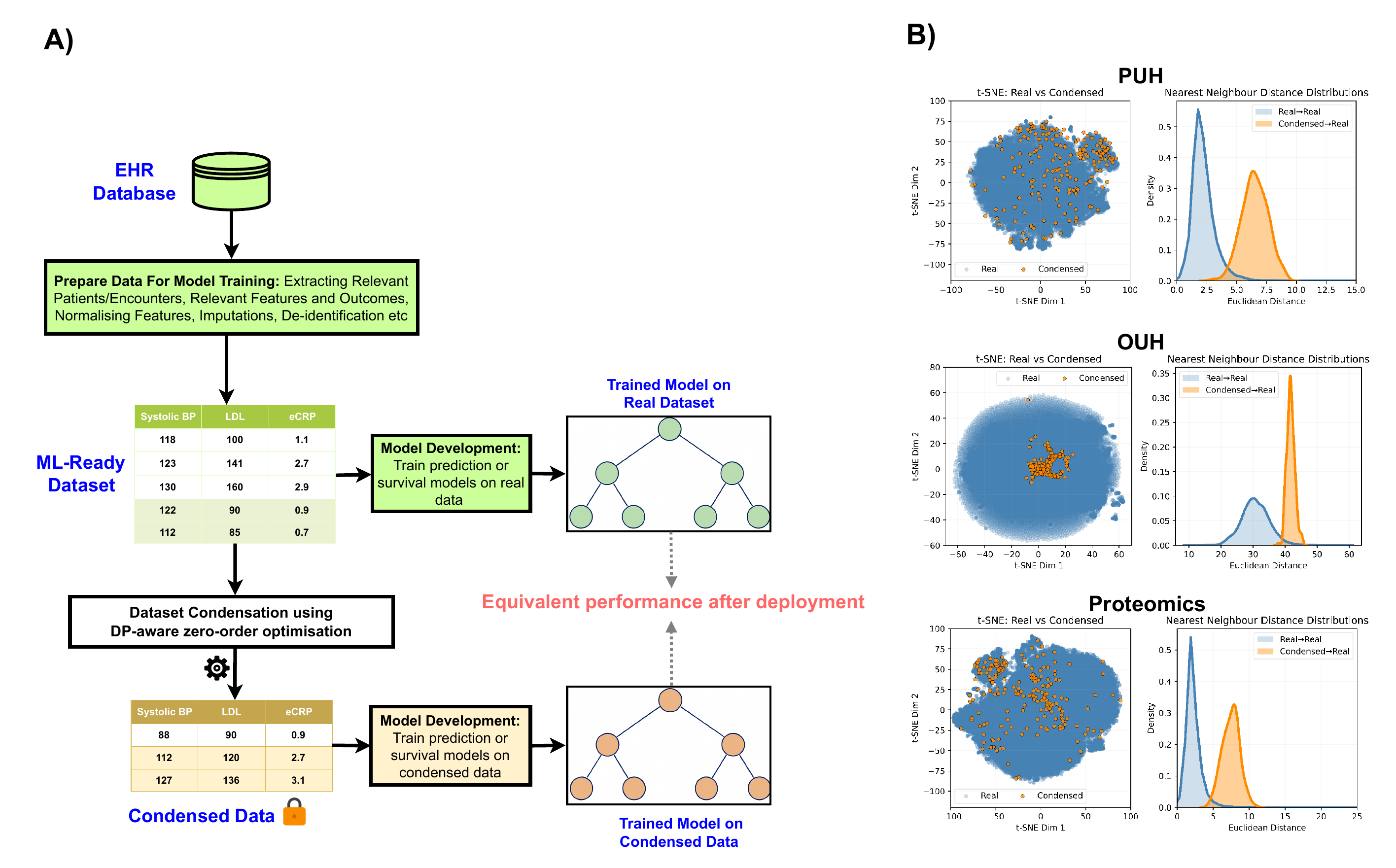}
\caption{Overview of dataset condensation (DC) workflow and structure of synthetic data. (A) Schematic showing how DC integrates into clinical ML pipelines. (B)  t-SNE projections (left) and nearest-neighbour distance distributions (right) for real and synthetic samples across PUH, OUH, and UK Biobank Proteomics datasets. Distributions compare distances from real to real and synthetic to real samples.}
\label{fig:tsne}
\end{figure}

\vspace{0.1cm}
\noindent While other privacy-preserving paradigms such as federated learning (FL) and generative modelling also aim to mitigate data access barriers, they differ fundamentally from DC in both design and utility. FL enables collaborative model training without centralising data, but it demands significant infrastructure and tightly coordinated participation across institutions \cite{mcmahan2017communication,rieke2020future,thakur2024knowledge}. Crucially, FL does not produce any reusable or transparent artifact such as a surrogate dataset that external researchers can access, examine, or use in downstream applications \cite{thakur2024data}. As a result, it offers no pathway to data democratisation and limits broader participation in clinical ML research. Generative models, such as generative adversarial networks (GANs) \cite{loni2025review,sun2025generating} or diffusion-based approaches \cite{zhong2024synthesizing}, attempt to recreate the full data distribution but often demand large training sets, extensive tuning, and prioritise sample realism over task-specific utility. In contrast, DC yields a small, shareable set of synthetic samples tailored for specific prediction tasks. By aggregating information across individual samples, DC reduces the risk of memorisation, avoids the need to replicate the full data distribution, and offers inherent privacy preservation as a by-product of its design. These properties make DC a uniquely practical and scalable approach for responsible data sharing in clinical ML.

\vspace{0.1cm}
\noindent Despite its growing appeal, DC remains poorly aligned with the modelling practices most common in clinical ML, where classical models such as decision trees, gradient-boosted ensembles, and Cox regression continue to dominate \cite{payne2020development,rajula2020comparison,shwartz2022tabular}. Most existing DC methods have been developed exclusively for neural networks and rely on gradient-based procedures to optimise synthetic samples \cite{yu2023dataset}. Many classical models, however, lack differentiability with respect to their parameters, making them incompatible with these approaches. In addition, differences in inductive biases between classical models and neural networks may cause condensed data generated by neural-network-based methods to generalise poorly when applied to classical models. This disparity highlights a critical gap between recent advances in DC and their practical integration into clinical ML workflows, which continue to rely heavily on classical approaches valued for their interpretability, regulatory familiarity, and proven performance on structured health data \cite{shwartz2022tabular,li2023regulating}.

\vspace{0.1cm}
\noindent To close this gap, we introduce a DC framework tailored to the classical models that remain central to clinical ML. The proposed approach begins by training a reference model on the real dataset, which is then treated as a \emph{black box}: we do not access its internal parameters or gradients, but only query its predictions. We then use a zero-order optimisation strategy---one that estimates how the model’s outputs change in response to small perturbations in the synthetic inputs, without requiring any differentiability \cite{zero}---to iteratively refine a compact pool of synthetic samples that act as surrogates for real patient records. These synthetic inputs are optimised so that the reference model's predictions align with both their assigned labels and the predictive distribution of the real dataset. This alignment allows models trained on synthetic data to faithfully reproduce the predictive patterns learned from real patient records. To ensure that no sensitive information from the real dataset is embedded in the condensed set, each update step is performed under calibrated Gaussian noise, yielding formal \((\varepsilon,\delta)\)-differential privacy guarantees for the original training data \cite{dwork2014algorithmic,abadi2016deep}. The result is a privacy-preserving condensed dataset that retains the clinical signal necessary for predictive modelling tasks such as diagnosis, risk stratification, and survival analysis, while avoiding the inclusion of sensitive patient-level information.

\vspace{0.1cm}
\noindent Figure~\ref{fig:tsne} illustrates how the proposed DC framework integrates into a typical clinical ML workflow and the nature of the synthetic data it produces. Panel A provides a schematic view of the broader context: DC operates on ML-ready datasets, typically derived from EHRs through cohort selection, feature extraction, and de-identification---whether by pseudonymisation (the most common practice) or full anonymisation---under institutionally approved governance \cite{vokinger2020lost,el2015anonymising}. Rather than replacing these safeguards, the proposed approach builds upon them, offering an additional layer of protection before model development. To examine the nature of the synthetic data generated by our method, Panel B shows a t-SNE projection \cite{van2008visualizing} of both real and synthetic samples across three datasets. In one of the datasets, synthetic samples tend to lie near the centre of the t-SNE space, while in others they are more broadly distributed across the full spread of the real data. This illustrates that the synthetic data can assume different configurations depending on the underlying structure, supporting utility across diverse datasets and tasks. Across all datasets, synthetic samples do not collapse onto specific real examples, as confirmed by nearest-neighbour distance plots showing greater separation from real data than among real samples themselves, indicating abstraction rather than memorisation. Together, these observations, along with the experimental evaluation presented in the remainder of this paper, highlight the potential of our framework as a practical and scalable solution for privacy-preserving data democratisation in clinical ML.

\begin{table}[h]
\centering
\caption{Summary of datasets used for evaluation, including sample sizes and associated prediction tasks. The SEER dataset refers to a curated breast cancer subset commonly used in survival modeling.}
\label{tab:data}

\begin{sc}
\begin{small}
\begin{tabular}{c|c|c}
\toprule
\textbf{Datasets}                     & \textbf{Samples} & \textbf{Task}                       \\ \toprule
Portsmouth University Hospitals (PUH) & 38,717           & \multirow{3}{*}{COVID-19 Prediction} \\
Oxford University Hospitals (OUH)     & 161,955          &                                     \\ 
University Hospitals Birmingham (UHB) & 95,236           &                                     \\ \midrule
UK Biobank –- Proteomics         & 52,999           & Myeloma Prediction                  \\ \midrule
SEER (Breast Cancer Subset)           & 4,024            & Cancer Survival Analysis            \\ \midrule
UK Biobank -– Diabetes                 & 474,886          & Diabetes Survival Analysis          \\ \bottomrule
\end{tabular}
\end{small}
\end{sc}

\end{table}

\section{Results}

\subsection{Datasets}
The performance of the proposed framework is evaluated on three CURIAL datasets \cite{soltan1,soltan2024scalable}, comprising anonymised EHRs that include demographic information, blood test results, and vital signs. These data were collected from emergency departments at three NHS Trusts: Portsmouth University Hospitals (PUH), Oxford University Hospitals (OUH), and University Hospitals Birmingham (UHB). These datasets supports a binary classification task: predicting whether a patient is COVID-19 positive. The framework is also assessed on high dimensional plasma proteomics data from the UK Biobank \cite{sudlow2015uk,sun2023plasma} for the prediction of incident multiple myeloma \cite{fieggen2025dysregulated}. These data represent scaled and normalised plasma concentrations for 2932 different proteins.

\vspace{0.2cm}
\noindent The framework is also evaluated on two datasets for survival prediction tasks. The first is a curated subset of the SEER cancer registry, focused on breast cancer patients \cite{teng2019seer}. This dataset includes anonymised records with variables such as age, sex, cancer stage, and follow-up time, and supports a time-to-event analysis task: predicting overall survival from the time of diagnosis. The second dataset is also derived from the UK Biobank \cite{sudlow2015uk} and supports a survival task focused on the future onset of diabetes. The objective is to model the time until individuals develop diabetes after baseline, using demographic, clinical, and lifestyle features collected at enrolment into the UK Biobank. In both datasets, right-censoring is appropriately accounted for, and all features are preprocessed through standardisation and encoding to ensure compatibility with survival models.

\vspace{0.2cm}
\noindent Table \ref{tab:data} provides a summary of these datasets, including details the number of samples and nature of tasks. In all datasets, 70\%, 10\%, and 20\% of the available examples are designated for training, validation, and testing, respectively. Additional information about the available features in each dataset are provided in Supplementary Note 1.

\begin{table}[!t]
\centering
\caption{Evaluation of models trained on condensed data for COVID-19 prediction using datasets from three NHS Trusts (a-c) and for myeloma prediction using UK Biobank plasma proteomics data (d). Results are reported as mean $\pm$ 95\% confidence intervals. Privacy budgets ($\epsilon$) are computed under $(\epsilon, \delta)$-differential privacy with $\delta = 10^{-5}$, where the privacy budget $\epsilon$ quantifies the trade-off between privacy and utility; smaller values of $\epsilon$ provide stronger privacy but may reduce data utility.}
\label{tab:res_pred}

\textbf{(a) Portsmouth University Hospitals (PUH)}\\[0.5ex]
\begin{sc}
\resizebox{0.72\linewidth}{!}{
\begin{tabular}{c|ccccc|c}
\toprule
\multirow{2}{*}{\textbf{Instances}} & \multicolumn{5}{c|}{\textbf{Metrics}} & \multirow{2}{*}{\shortstack{\textbf{Privacy}\\ \textbf{Budget} \\ \textbf{($\epsilon$)}}} \\ \cmidrule{2-6}
& AUROC & Sensitivity & Specificity & PPV & NPV & \\
\toprule
50   & 0.862 $\pm$ 0.023 & 0.733 $\pm$ 0.042 & 0.839 $\pm$ 0.009 & 0.205 $\pm$ 0.013 & 0.982 $\pm$ 0.003 & 2.5 \\
100  & 0.894 $\pm$ 0.021 & 0.756 $\pm$ 0.044 & 0.887 $\pm$ 0.008 & 0.286 $\pm$ 0.018 & 0.985 $\pm$ 0.003 & 2.6 \\
500  & 0.888 $\pm$ 0.022 & \textbf{0.782} $\pm$ 0.042 & 0.871 $\pm$ 0.008 & 0.271 $\pm$ 0.017 & \textbf{0.986} $\pm$ 0.003 & 1.8 \\
1000 & 0.884 $\pm$ 0.022 & 0.741 $\pm$ 0.041 & 0.868 $\pm$ 0.007 & 0.262 $\pm$ 0.016 & 0.983 $\pm$ 0.003 & 3.1 \\
\midrule
\emph{\textbf{Full Dataset}} & \textbf{0.901} $\pm$ 0.022 & 0.751 $\pm$ 0.045 & \textbf{0.905} $\pm$ 0.006 & \textbf{0.316} $\pm$ 0.026 & 0.985 $\pm$ 0.003 & -- \\
\bottomrule
\end{tabular}
}
\end{sc}

\vspace{3ex}
\textbf{(b) Oxford University Hospitals (OUH)}\\[0.5ex]
\begin{sc}
\resizebox{0.72\linewidth}{!}{
\begin{tabular}{c|ccccc|c}
\toprule
\multirow{2}{*}{\textbf{Instances}} & \multicolumn{5}{c|}{\textbf{Metrics}} & \multirow{2}{*}{\shortstack{\textbf{Privacy}\\ \textbf{Budget} \\ \textbf{($\epsilon$)}}} \\ \cmidrule{2-6}
& AUROC & Sensitivity & Specificity & PPV & NPV & \\
\toprule
50   & 0.855 $\pm$ 0.019 & 0.752 $\pm$ 0.037 & 0.819 $\pm$ 0.007 & 0.067 $\pm$ 0.004 & 0.995 $\pm$ 0.001 & 2.2 \\
100  & 0.873 $\pm$ 0.019 & 0.744 $\pm$ 0.038 & 0.847 $\pm$ 0.003 & 0.077 $\pm$ 0.004 & 0.995 $\pm$ 0.001 & 2.7 \\
500  & 0.870 $\pm$ 0.019 & 0.725 $\pm$ 0.039 & 0.843 $\pm$ 0.004 & 0.073 $\pm$ 0.004 & 0.994 $\pm$ 0.001 & 2.1 \\
1000 & 0.891 $\pm$ 0.018 & \textbf{0.760} $\pm$ 0.038 & 0.887 $\pm$ 0.005 & 0.095 $\pm$ 0.005 & 0.995 $\pm$ 0.001 & 1.9 \\
\midrule
\emph{\textbf{Full Dataset}} & \textbf{0.911} $\pm$ 0.018 & 0.756 $\pm$ 0.035 & \textbf{0.898} $\pm$ 0.003 & \textbf{0.126} $\pm$ 0.008 & \textbf{0.996} $\pm$ 0.001 & -- \\
\bottomrule
\end{tabular}
}
\end{sc}

\vspace{3ex}
\textbf{(c) University Hospitals Birmingham (UHB)}\\[0.5ex]
\begin{sc}
\resizebox{0.72\linewidth}{!}{
\begin{tabular}{c|ccccc|c}
\toprule
\multirow{2}{*}{\textbf{Instances}} & \multicolumn{5}{c|}{\textbf{Metrics}} & \multirow{2}{*}{\shortstack{\textbf{Privacy}\\ \textbf{Budget} \\ \textbf{($\epsilon$)}}} \\
\cmidrule{2-6}
 & AUROC & Sensitivity & Specificity & PPV & NPV & \\
\toprule
50   & 0.849 $\pm$ 0.037 & 0.685 $\pm$ 0.074 & 0.825 $\pm$ 0.006 & 0.032 $\pm$ 0.004 & 0.997 $\pm$ 0.001 & 1.99 \\
100  & 0.909 $\pm$ 0.030 & 0.826 $\pm$ 0.057 & 0.849 $\pm$ 0.005 & 0.074 $\pm$ 0.003 & 0.998 $\pm$ 0.001 & 2.20 \\
500  & 0.889 $\pm$ 0.033 & 0.799 $\pm$ 0.060 & 0.840 $\pm$ 0.005 & 0.069 $\pm$ 0.005 & 0.998 $\pm$ 0.001 & 3.19 \\
1000 & 0.886 $\pm$ 0.033 & 0.795 $\pm$ 0.052 & 0.838 $\pm$ 0.006 & 0.07 $\pm$ 0.005 & 0.998 $\pm$ 0.001 & 3.45 \\
\midrule
\emph{\textbf{Full Dataset}} & \textbf{0.925} $\pm$ 0.036 & \textbf{0.745} $\pm$ 0.070 & \textbf{0.936} $\pm$ 0.003 & \textbf{0.135} $\pm$ 0.016 & \textbf{0.998} $\pm$ 0.001 & -- \\
\bottomrule
\end{tabular}}
\end{sc}

\vspace{3ex}
\textbf{(d) Proteomic (UK Biobank)}\\[0.5ex]
\begin{sc}
\resizebox{0.72\linewidth}{!}{
\begin{tabular}{c|ccccc|c}
\toprule
\multirow{2}{*}{\textbf{Instances}} & \multicolumn{5}{c|}{\textbf{Metrics}} &

\multirow{2}{*}{\shortstack{\textbf{Privacy}\\\textbf{Budget} \\ \textbf{($\epsilon$)}}}

\\ \cmidrule{2-6}
 & AUROC & Sensitivity & Specificity & PPV & NPV & \\
\toprule
50  & 0.856 $\pm$ 0.053 & 0.788 $\pm$ 0.106 & 0.817 $\pm$ 0.003 & 0.014 $\pm$ 0.002 & 0.999 $\pm$ 0.001 & 2.1 \\
100 & 0.905 $\pm$ 0.048 & 0.846 $\pm$ 0.093 & 0.785 $\pm$ 0.006 & 0.013 $\pm$ 0.001 & 0.999 $\pm$ 0.001 & 2.6 \\
500 & \textbf{0.913} $\pm$ 0.044 & \textbf{0.865} $\pm$ 0.092 & 0.834 $\pm$ 0.006 & 0.017 $\pm$ 0.002 & 0.999 $\pm$ 0.001 & 1.9 \\
1000 & 0.912 $\pm$ 0.052 & 0.808 $\pm$ 0.096 & \textbf{0.879} $\pm$ 0.005 & \textbf{0.022} $\pm$ 0.003 & 0.999 $\pm$ 0.001 & 2.3 \\
\midrule
\emph{\textbf{Full Dataset}} & 0.898 $\pm$ 0.050 & 0.846 $\pm$ 0.091 & 0.757 $\pm$ 0.006 & 0.011 $\pm$ 0.001 & \textbf{0.999} $\pm$ 0.001 & -- \\
\bottomrule
\end{tabular}}
\end{sc}

\end{table}

\subsection{Performance on Prediction Tasks}

We trained gradient-boosted decision tree models (XGBoost) on the CURIAL and UK Biobank proteomics datasets to guide the optimisation of their corresponding condensed datasets. Condensed data were learned under varying instance-per-class (IPC) settings, specifically IPC 50, 100, 500, and 1000, where IPC denotes the number of synthetic examples generated per class. These condensed datasets were then used to train new XGBoost models, which were evaluated on held-out test data to assess predictive performance and privacy guarantees. Table \ref{tab:res_pred} summarises model performance across metrics for each IPC setting.

\vspace{0.1cm}
\noindent As shown in Table \ref{tab:res_pred}, models trained on condensed data achieved strong predictive performance across all CURIAL datasets: PUH, OUH, and UHB. The best AUROC scores were obtained at IPC 100 for PUH (0.894) and UHB (0.909), closely matching their respective full-data baselines of 0.901 and 0.925. At OUH, the highest AUROC (0.891) was achieved at IPC 1000, also approaching the full-data score of 0.911. These results demonstrate that condensed datasets with just 100 instances per class can achieve near-parity with full datasets, retaining discriminative power with far fewer examples. Sensitivity was also well-preserved across IPC levels, and in some cases even exceeded that of the full-data models, for example PUH at IPC 500 (0.782 vs. 0.751). These gains were achieved without substantial losses in specificity, indicating that the condensed data preserved discriminative thresholds effectively. 

\vspace{0.1cm}
\noindent Positive predictive values (PPV) remained modest across all sites due to the underlying class imbalance in COVID-19 diagnosis. However, negative predictive values (NPV) consistently exceeded 0.98, indicating reliable identification of true negatives. For example, at IPC 100, NPV values were 0.985 (PUH), 0.995 (OUH), and 0.998 (UHB), nearly identical to those obtained using the full datasets. This suggests that models trained on condensed data are particularly well-suited for screening scenarios where safe and confident exclusion of disease is clinically critical.

\vspace{0.1cm}
\noindent For the proteomics dataset, which involves predicting future multiple myeloma from baseline plasma protein profiles, models trained on condensed data achieved performance that matched or exceeded models trained on the full dataset. The highest AUROC was 0.913 at IPC 500, surpassing the full-data baseline of 0.898. IPC 100 and IPC 1000 also showed strong performance, with AUROCs of 0.905 and 0.912. These results indicate that the condensed data can effectively retain predictive signal from complex molecular data. Positive predictive value (PPV) was consistently higher for condensed models compared to the full-data baseline. Negative predictive value (NPV) remained above $0.99$ across all settings. Together, these results suggest that condensation not only preserves overall model discrimination but also improves the precision and clinical utility of predictions. This improvement may be partly explained by the limited number of positive cases in the original training data, which included only 98 individuals who developed multiple myeloma. In this context, generating more condensed examples for the minority class likely acted as an implicit form of data augmentation, helping the model generalise more effectively to unseen cases.

\vspace{0.1cm}
\noindent In addition to predictive utility, the proposed method provides formal privacy guarantees through differential privacy, which bounds the influence of individual training examples and limits susceptibility to membership inference attacks. Strong performance was retained even under moderate privacy budgets. For the CURIAL datasets, the best-performing IPC setting was IPC 100 for both PUH ($\varepsilon = 2.6$) and UHB ($\varepsilon = 2.2$), and IPC 1000 for OUH ($\varepsilon = 1.9$), all with $\delta = 10^{-5}$. In the proteomics task, the best AUROC was observed at IPC 500 with a privacy budget of $\varepsilon = 1.9$. These values fall within a range considered to offer effective privacy protection in practical settings \cite{abadi2016deep,kaissis2020secure}. Together, these results demonstrate that the proposed condensation framework supports high model utility while maintaining quantifiable and meaningful privacy guarantees.

\begin{table}[!t]
\centering
\caption{Evaluation of survival model performance (C-index with 95\% CI) and privacy budget ($\varepsilon$) across IPC levels over the UK Biobank diabetes and SEER datasets. $\delta$ is fixed at $10^{-5}$.}
\label{tab:res_surv}
\begin{sc}
\begin{tabular}{llccccc}
\multicolumn{7}{c}{\textbf{Dataset: Diabetes (UK Biobank)}} \\
\toprule
\textbf{Model} & \textbf{Metric} & \textbf{Full} & \textbf{IPC 50} & \textbf{IPC 100} & \textbf{IPC 500} & \textbf{IPC 1000} \\
\midrule
\multirow{2}{*}{COX} & C-index & \textbf{0.79} $\pm$ 0.002 & 0.787 $\pm$ 0.004 & \textbf{0.79} $\pm$ 0.003 & 0.788 $\pm$ 0.002 & 0.789 $\pm$ 0.003 \\
                     & $\varepsilon$ & — & 1.91 & 1.937 & 2.442 & 2.54 \\ \cmidrule{2-7}
\multirow{2}{*}{XGB} & C-index & \textbf{0.802} $\pm$ 0.003 & 0.762 $\pm$ 0.004 & 0.775 $\pm$ 0.003 & 0.786 $\pm$ 0.004 & 0.788 $\pm$ 0.002 \\
                     & $\varepsilon$ & — & 0.491 & 0.560 & 0.736 & 1.146 \\ \midrule
\addlinespace[1em]
\multicolumn{7}{c}{\textbf{Dataset: SEER}} \\
\toprule
\textbf{Model} & \textbf{Metric} & \textbf{Full} & \textbf{IPC 50} & \textbf{IPC 100} & \textbf{IPC 500} & \textbf{IPC 1000} \\
\midrule
\multirow{2}{*}{COX} & C-index & 0.729 $\pm$ 0.028 & 0.694 $\pm$ 0.025 & 0.714 $\pm$ 0.021 & \textbf{0.732} $\pm$ 0.027 & 0.724 $\pm$ 0.027 \\
                     & $\varepsilon$ & — & 1.801 & 2.611 & 2.920 & 3.789 \\ \cmidrule{2-7}
\multirow{2}{*}{XGB} & C-index & \textbf{0.731} $\pm$ 0.024 & 0.689 $\pm$ 0.029 & 0.716 $\pm$ 0.026 & 0.725 $\pm$ 0.022 & 0.724 $\pm$ 0.027 \\
                     & $\varepsilon$ & — & 1.413 & 2.275 & 1.916 & 2.317 \\
\bottomrule
\end{tabular}
\end{sc}
\end{table}

\subsection{Performance on Survival Analysis Tasks}
Similar patterns were observed in the survival analysis tasks, where models trained on condensed data achieved performance comparable to those trained on the full datasets. For each dataset and model, we used the same model class for both condensation and downstream evaluation. Specifically, condensed datasets were optimised using either a Cox proportional hazards model \cite{cox1972regression} or the Accelerated Failure Time (AFT) variant of XGBoost \cite{barnwal2022survival}, depending on the target model used for deployment. As summarised in Table~\ref{tab:res_surv}, performance remained stable across IPC levels, with several configurations achieving near-parity with or even exceeding full-data baselines.

\vspace{0.1cm}
\noindent For the Diabetes dataset from UK Biobank, the Cox model reached its best performance at IPC 100 with a C-index of 0.79, identical to the full-data model. The XGBoost model improved steadily across IPC levels, achieving a C-index of 0.788 at IPC 1000 compared to 0.802 on the full dataset. On the SEER dataset, the Cox model performed best at IPC 500 with a C-index of 0.732, slightly exceeding the full-data baseline of 0.729. The XGBoost model achieved comparable results at IPC 500 and IPC 1000 relative to the full-data performance. These results indicate that the condensation framework supports reliable time-to-event modelling across datasets and architectures.

\begin{figure}[!t]
\centering
\includegraphics[scale=0.65, trim=20pt 0pt 10pt 10pt, clip]{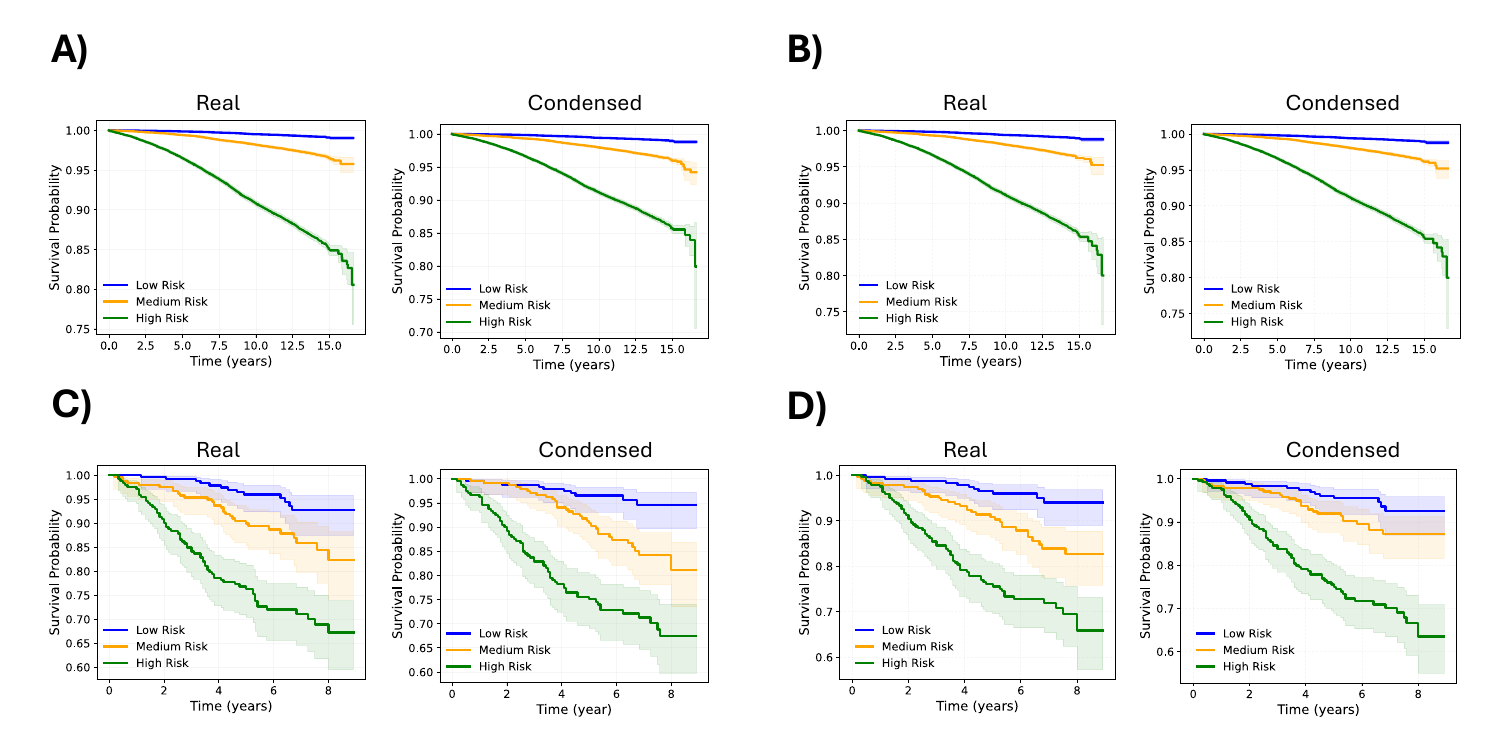}
\caption{Kaplan–Meier (KM) curves for models trained on real and condensed datasets. The first row shows KM curves from \textbf{A)} XGBoost and \textbf{B)} Cox models trained on the Diabetes (UK Biobank) dataset and their respective best-performing condensed datasets. The second row shows corresponding KM curves from \textbf{C)} XGBoost and \textbf{D)} Cox models trained on the SEER dataset.}

\label{fig:KM}
\end{figure}

\vspace{0.1cm}
\noindent To further assess utility, we compared survival distributions produced by models trained on full versus condensed data. As shown in Figure~\ref{fig:KM}, Kaplan-Meier (KM) curves \cite{klein2003survival} were closely aligned across all tasks. The first row shows KM curves for XGBoost and Cox models trained on the Diabetes dataset, while the second row shows the corresponding results for the SEER dataset. In all cases, the curves derived from condensed data mirrored those from full data, suggesting that the synthetic samples retained survival-relevant structure effectively.

\vspace{0.1cm}
\noindent \noindent These strong utility results were achieved under meaningful privacy guarantees. For XGBoost on the Diabetes dataset, performance remained competitive under strong differential privacy constraints, with privacy budgets as low as $\varepsilon = 0.560$ at IPC 100 and $\varepsilon = 1.146$ at IPC 1000. For Cox models, the best-performing condensed datasets were obtained with $\varepsilon = 1.937$ (Diabetes) and $\varepsilon = 2.920$ (SEER), offering effective privacy protection while preserving model accuracy.

\subsection{Generalisation to External Cohorts and Models}
The proposed framework generates condensed datasets by training solely on a specific dataset and a corresponding base model (e.g., XGBoost).While the primary use case assumes deployment within the same data distribution and model type, it is valuable to examine the broader utility of the condensed data. Specifically, we evaluate whether (1) models trained on condensed data can maintain performance when tested on external cohorts, and (2) the condensed data can support effective training of other downstream models not used during condensation. These properties align with the intended use of condensed datasets as portable, model-agnostic surrogates for real clinical datasets.

\vspace{0.2cm}
\noindent \textbf{External validation:} To assess the generalisation of models trained on condensed datasets, we conduct cross-site external validation using the three CURIAL datasets. Specifically, models trained on condensed data from PUH and OUH are evaluated on the UHB test set, while those trained on PUH and UHB are evaluated on the OUH test set. This setup allows us to examine how well the condensed data capture generalisable patterns across clinical populations. Table~\ref{tab:ev} presents the results of this evaluation.

\vspace{0.1cm}
\noindent On the UHB test set, models trained on condensed data from both PUH and OUH consistently outperform those trained on the corresponding real datasets. For example, AUROC improves from 0.830 to 0.861 in the PUH case, and from 0.842 to 0.864 in the OUH case indicating stronger discriminatory performance. These gains are accompanied by substantial increases in sensitivity, while specificity, NPV, and PPV are maintained or modestly improved. Similar trends are observed on the OUH test set, where models trained on condensed data again outperform their full-data counterparts.

\vspace{0.1cm}
\noindent One possible explanation for this behaviour is that the condensation process acts as an effective form of regularisation---potentially selecting and amplifying the most generic patterns in the data while suppressing site-specific characteristics and spurious correlations. In our setting, this may help models trained on condensed data learn more robust and generalisable decision boundaries, leading to improved performance on external cohorts.

\begin{table}[!t]
\centering
\caption{Cross-site generalisation performance across four train--test pairs. Each row shows the performance of models trained on condensed datasets generated from one hospital site and evaluated on a held-out test set from another site. The first row reports results on the UHB test set using models trained on data condensed from PUH and OUH. The second row reports results on the OUH test set using models trained on data condensed from PUH and UHB.}
\label{tab:ev}
\begin{sc}
\resizebox{!}{0.065\textwidth}{
\begin{tabular}{l!{\vrule width 0.75pt}ccccc!{\vrule width 0.75pt}ccccc}
\toprule
\textbf{Model} 
& \multicolumn{5}{c!{\vrule width 0.75pt}}{\textbf{PUH $\rightarrow$ UHB}} 
& \multicolumn{5}{c}{\textbf{OUH $\rightarrow$ UHB}} \\
\cmidrule(lr){2-6} \cmidrule(lr){7-11}
& AUROC & Sensitivity & Specificity & PPV & NPV 
& AUROC & Sensitivity & Specificity & PPV & NPV \\
\midrule
Full      
& 0.830 $\pm$ 0.037 & 0.631 $\pm$ 0.074 & \textbf{0.866} $\pm$ 0.005 & 0.038 $\pm$ 0.005 & 0.996 $\pm$ 0.001 
& 0.842 $\pm$ 0.040 & 0.611 $\pm$ 0.077 & \textbf{0.917} $\pm$ 0.004 & \textbf{0.058} $\pm$ 0.008 & 0.996 $\pm$ 0.001 \\
IPC 50    
& 0.855 $\pm$ 0.033 & 0.758 $\pm$ 0.064 & 0.847 $\pm$ 0.006 & 0.040 $\pm$ 0.004 & \textbf{0.998} $\pm$ 0.001 
& 0.825 $\pm$ 0.037 & 0.711 $\pm$ 0.074 & 0.814 $\pm$ 0.006 & 0.031 $\pm$ 0.003 & 0.997 $\pm$ 0.001 \\
IPC 100   
& 0.860 $\pm$ 0.035 & 0.765 $\pm$ 0.070 & 0.840 $\pm$ 0.004 & 0.039 $\pm$ 0.003 & \textbf{0.998} $\pm$ 0.001 
& 0.831 $\pm$ 0.035 & 0.691 $\pm$ 0.074 & 0.863 $\pm$ 0.005 & 0.041 $\pm$ 0.004 & 0.997 $\pm$ 0.001 \\
IPC 500   
& 0.847 $\pm$ 0.038 & 0.698 $\pm$ 0.074 & 0.836 $\pm$ 0.003 & 0.035 $\pm$ 0.004 & 0.997 $\pm$ 0.001 
& \textbf{0.864} $\pm$ 0.033 & \textbf{0.772} $\pm$ 0.064 & 0.833 $\pm$ 0.006 & 0.037 $\pm$ 0.003 & \textbf{0.998} $\pm$ 0.001 \\
IPC 1000  
& \textbf{0.861} $\pm$ 0.032 & \textbf{0.805} $\pm$ 0.064 & 0.845 $\pm$ 0.005 & \textbf{0.042} $\pm$ 0.003 & \textbf{0.998} $\pm$ 0.001 
& 0.844 $\pm$ 0.035 & 0.745 $\pm$ 0.070 & 0.862 $\pm$ 0.005 & \textbf{0.043} $\pm$ 0.004 & \textbf{0.998} $\pm$ 0.001 \\
\bottomrule
\end{tabular}}
\end{sc}
\vspace{2em} 

\begin{sc}
\resizebox{!}{0.065\textwidth}{
\begin{tabular}{l|ccccc|ccccc}
\toprule
\textbf{Model} 
& \multicolumn{5}{c|}{\textbf{PUH $\rightarrow$ OUH}} 
& \multicolumn{5}{c}{\textbf{UHB $\rightarrow$ OUH}} \\
\cmidrule(lr){2-6} \cmidrule(lr){7-11}
& AUROC & Sensitivity & Specificity & PPV & NPV 
& AUROC & Sensitivity & Specificity & PPV & NPV \\
\midrule
Full       
& 0.863 $\pm$ 0.021 & 0.705 $\pm$ 0.041 & \textbf{0.885} $\pm$ 0.003 & \textbf{0.095} $\pm$ 0.006 & 0.994 $\pm$ 0.001 
& 0.778 $\pm$ 0.020 & 0.326 $\pm$ 0.041 & \textbf{0.953} $\pm$ 0.002 & \textbf{0.085} $\pm$ 0.013 & 0.988 $\pm$ 0.001 \\
IPC 60     
& 0.862 $\pm$ 0.019 & 0.723 $\pm$ 0.039 & 0.844 $\pm$ 0.004 & 0.074 $\pm$ 0.004 & 0.994 $\pm$ 0.001 
& 0.806 $\pm$ 0.022 & 0.502 $\pm$ 0.043 & 0.891 $\pm$ 0.003 & 0.070 $\pm$ 0.006 & 0.991 $\pm$ 0.001 \\
IPC 100    
& \textbf{0.873} $\pm$ 0.019 & 0.721 $\pm$ 0.039 & 0.874 $\pm$ 0.004 & 0.089 $\pm$ 0.005 & \textbf{0.995} $\pm$ 0.001 
& \textbf{0.821} $\pm$ 0.021 & \textbf{0.613} $\pm$ 0.040 & 0.870 $\pm$ 0.004 & 0.072 $\pm$ 0.005 & \textbf{0.993} $\pm$ 0.001 \\
IPC 500    
& 0.848 $\pm$ 0.020 & 0.727 $\pm$ 0.039 & 0.823 $\pm$ 0.004 & 0.066 $\pm$ 0.004 & 0.994 $\pm$ 0.001 
& 0.806 $\pm$ 0.022 & 0.525 $\pm$ 0.041 & 0.886 $\pm$ 0.003 & 0.070 $\pm$ 0.005 & 0.991 $\pm$ 0.001 \\
IPC 1000   
& 0.851 $\pm$ 0.020 & \textbf{0.729} $\pm$ 0.039 & 0.848 $\pm$ 0.004 & 0.076 $\pm$ 0.004 & \textbf{0.995} $\pm$ 0.001 
& 0.803 $\pm$ 0.022 & 0.517 $\pm$ 0.042 & 0.871 $\pm$ 0.004 & 0.062 $\pm$ 0.005 & \textbf{0.991} $\pm$ 0.001 \\
\bottomrule
\end{tabular}}
\end{sc}
\end{table}

\begin{figure}[!t]
\centering
\includegraphics[scale=0.42, trim=12pt 0pt 10pt 10pt, clip]{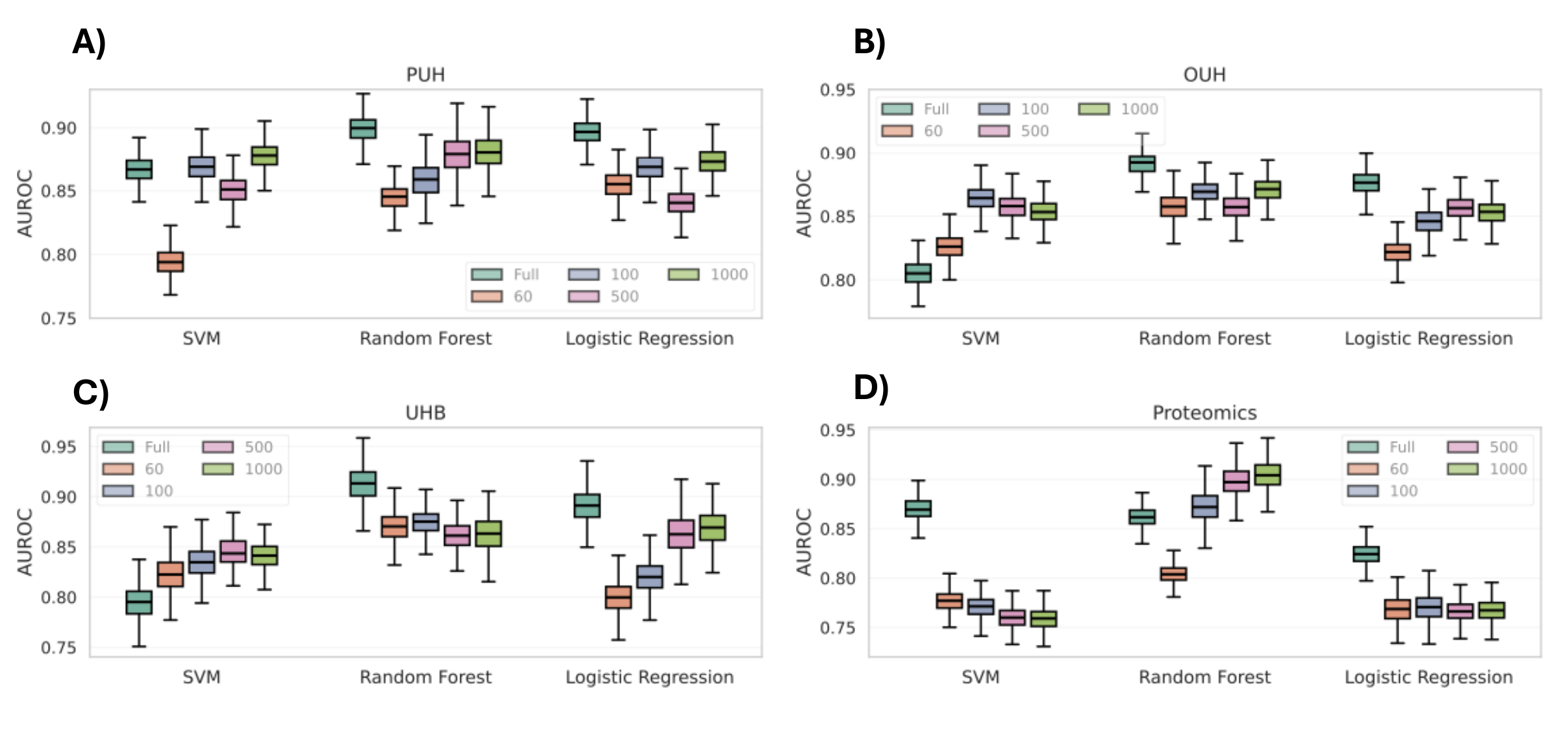}
\caption{Cross-model evaluation of XGBoost-derived condensed data. Performance of support vector machine (SVM), Random Forest, and Logistic Regression models trained on condensed data generated using XGBoost, across four datasets: (A) PUH, (B) OUH, (C) UHB, and (D) Proteomics.}
\label{fig:models}
\end{figure}

\vspace{0.2cm}
\noindent \textbf{Generalisation to other models:} We trained support vector machines (SVM), random forests, and logistic regression models on the condensed datasets and evaluated their performance. This analysis assesses whether the condensed data preserves task-relevant structure beyond the model used during condensation. Figure~\ref{fig:models} illustrates the results of this experiment. Across the three CURIAL sites (PUH, OUH, and UHB), we observe that models trained on condensed data, originally generated using XGBoost, generalise well to other classifiers. In most cases, both random forests and logistic regression trained on condensed data achieve comparable performance levels to those trained on the full dataset, highlighting the utility of the condensed data across model types. 

\vspace{0.1cm}
\noindent Notably, SVM classifiers trained on condensed data outperform their full-data counterparts at PUH and OUH. This counterintuitive result may be explained by the condensation process again acting as a form of implicit regularisation which removes noisy, site-specific patterns and amplifies the most discriminative features. Since SVMs are particularly sensitive to margin noise and decision boundary overlap \cite{manwani2013noise}, they appear to benefit from this simplification, resulting in improved generalisation.

\begin{table}[!t]
\centering
\caption{Top 15 covariates ranked by hazard ratio (HR) with 95\% confidence intervals from Cox models trained on real and condensed variants of (a) the Diabetes and (b) SEER datasets.}
\label{tab:hr}
\begin{subtable}{\textwidth}
\centering
\caption{Diabetes (UK Biobank)}
\begin{sc}
\begin{small}
\resizebox{!}{0.15\textwidth}{
\begin{tabular}{cc}
\textbf{Real Dataset} & \textbf{Synthetic Dataset} \\
\toprule
\begin{tabular}{lc}
\textbf{Covariate} & \textbf{HR [95\% CI]} \\
\midrule
BMI & 1.288 [1.278, 1.298] \\
Triglycerides & 1.171 [1.161, 1.181] \\
Prev. Hypertension & 1.140 [1.130, 1.150] \\
Age & 1.108 [1.098, 1.118] \\
Gamma-GT & 1.087 [1.082, 1.092] \\
Deprivation Index & 1.080 [1.070, 1.090] \\
Ethnicity: South Asian & 1.057 [1.047, 1.067] \\
Smoker: Current & 1.046 [1.036, 1.056] \\
Sex: Male & 1.039 [1.029, 1.049] \\
Ethnicity: Black & 1.035 [1.030, 1.040] \\
LDL & 0.963 [0.953, 0.973] \\
Sex: Female & 0.962 [0.952, 0.972] \\
Smoker: Never & 0.960 [0.950, 0.970] \\
IGF-1 & 0.949 [0.939, 0.959] \\
Ethnicity: White & 0.940 [0.930, 0.950] \\
\end{tabular}
&
\begin{tabular}{lc}
\textbf{Covariate} & \textbf{HR [95\% CI]} \\
\midrule
BMI & 1.526 [1.381, 1.671] \\
Triglycerides & 1.320 [1.215, 1.425] \\
Prev. Hypertension & 1.233 [1.123, 1.343] \\
Age & 1.223 [1.118, 1.328] \\
Gamma-GT & 1.209 [1.119, 1.299] \\
Deprivation Index & 1.116 [1.026, 1.206] \\
Ethnicity: South Asian & 1.089 [1.001, 1.177] \\
Ethnicity: Other & 1.069 [0.989, 1.149] \\
Ethnicity: Black & 1.061 [0.966, 1.156] \\
Smoker: Current & 1.053 [0.968, 1.138] \\
LDL & 0.974 [0.894, 1.054] \\
IGF-1 & 0.935 [0.860, 1.010] \\
Smoker: Never & 0.933 [0.858, 1.008] \\
Sex: Female & 0.917 [0.847, 0.987] \\
Ethnicity: White & 0.895 [0.815, 0.975] \\
\end{tabular}
\\
\bottomrule
\end{tabular}}
\end{small}
\end{sc}
\end{subtable}

\vspace{0.5cm}

\begin{subtable}{\textwidth}
\centering
\caption{SEER Dataset}
\begin{sc}
\begin{small}
\resizebox{!}{0.15\textwidth}{
\begin{tabular}{cc}
\textbf{Real Dataset} & \textbf{Synthetic Dataset} \\
\toprule
\begin{tabular}{lc}
\textbf{Covariate} & \textbf{HR [95\% CI]} \\
\midrule
Regional Nodes Positive & 1.155 [1.055, 1.255] \\
Grade III & 1.098 [1.003, 1.193] \\
N Stage N3 & 1.092 [0.987, 1.197] \\
6th Stage IIIC & 1.092 [0.987, 1.197] \\
Age & 1.088 [1.008, 1.168] \\
Estrogen Negative & 1.083 [0.988, 1.178] \\
Progesterone Negative & 1.078 [0.978, 1.178] \\
T Stage T4 & 1.077 [1.007, 1.147] \\
Race Black & 1.074 [0.989, 1.159] \\
Tumor Size & 1.072 [0.982, 1.162] \\
Grade I & 0.930 [0.850, 1.010] \\
Progesterone Positive & 0.928 [0.843, 1.013] \\
Estrogen Positive & 0.924 [0.839, 1.009] \\
N Stage N1 & 0.896 [0.806, 0.986] \\
Regional Nodes Examined & 0.892 [0.817, 0.967] \\
\end{tabular}
&
\begin{tabular}{lc}
\textbf{Covariate} & \textbf{HR [95\% CI]} \\
\midrule
Regional Nodes Positive & 1.134 [1.059, 1.209] \\
N Stage N3 & 1.104 [1.034, 1.174] \\
T Stage T4 & 1.099 [1.029, 1.169] \\
Estrogen Negative & 1.092 [1.017, 1.167] \\
Grade III & 1.086 [1.016, 1.156] \\
Age & 1.071 [1.006, 1.136] \\
Race Black & 1.061 [0.991, 1.131] \\
Race Other & 0.943 [0.878, 1.008] \\
Grade II & 0.941 [0.881, 1.001] \\
Estrogen Positive & 0.939 [0.879, 0.999] \\
Married & 0.927 [0.867, 0.987] \\
6th Stage IIA & 0.920 [0.860, 0.980] \\
T Stage T1 & 0.910 [0.850, 0.970] \\
Regional Nodes Examined & 0.903 [0.848, 0.958] \\
Progesterone Positive & 0.859 [0.804, 0.914] \\
\end{tabular}
\\
\bottomrule
\end{tabular}}
\end{small}
\end{sc}
\end{subtable}
\end{table}

\vspace{0.1cm}
\noindent For the proteomics dataset, the condensed data does not generalise well to SVM and logistic regression, which perform substantially worse than their full-data counterparts. This likely reflects a mismatch between the linear assumptions of these models and the complex, high-dimensional structure of proteomic features. In contrast, random forests, which share inductive biases with XGBoost---the model used during condensation---generalise effectively and often outperform models trained on full data. This suggests that while the condensed data is well-suited to tree-based classifiers, it may not preserve the global structure required by simpler linear models. This behaviour underscores the importance of aligning the condensation process with the intended downstream models.

\subsection{Interpretability Comparison of Real and Condensed Models}
\begin{figure}[!t]
\centering
\begin{subfigure}[b]{0.47\textwidth}
    \centering
    \includegraphics[scale=0.405]{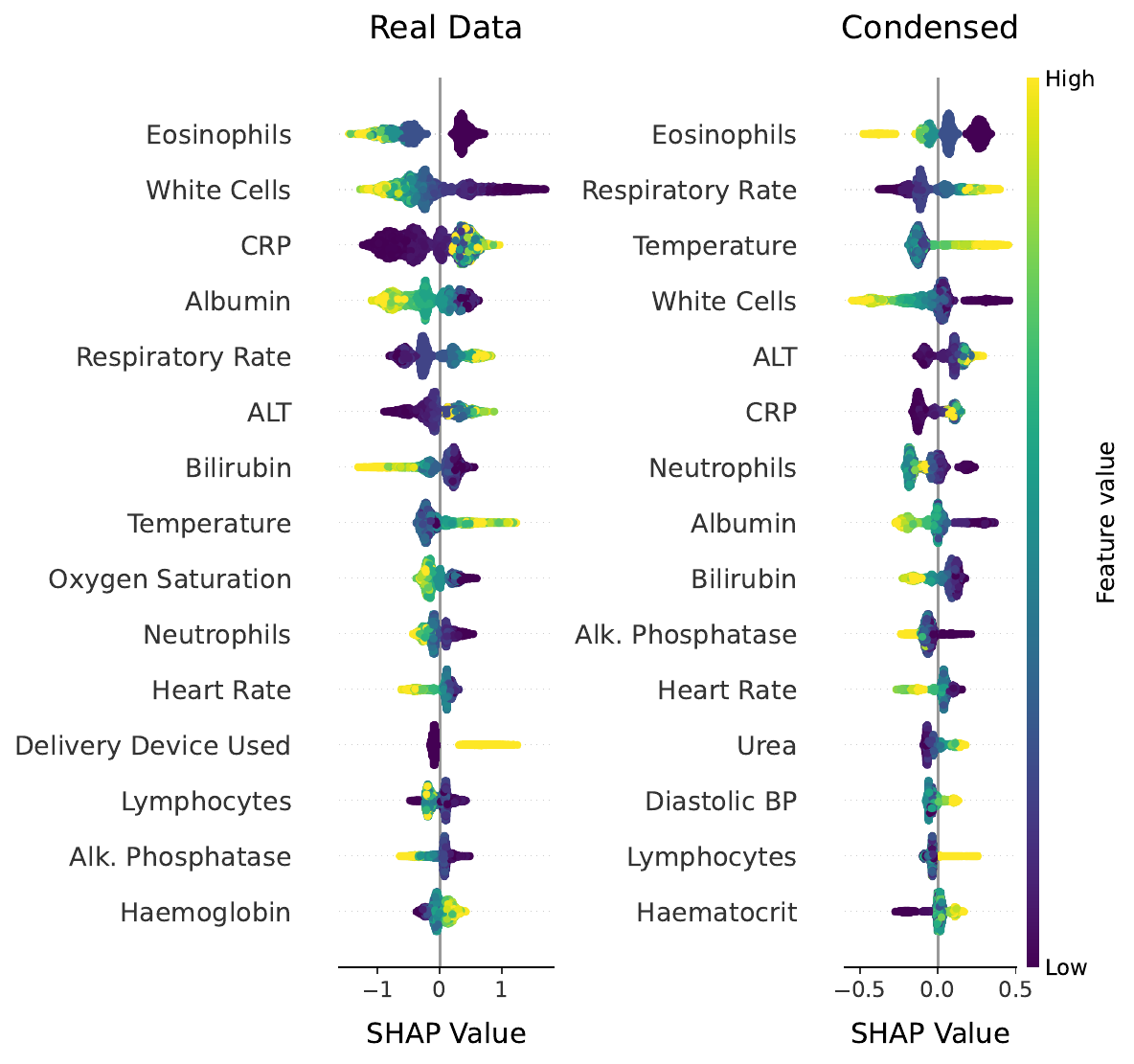}
    \caption{PUH dataset}
    \label{fig:shap_PUH}
\end{subfigure}
\hspace{0.02\textwidth}
\begin{subfigure}[b]{0.47\textwidth}
    \centering
    \includegraphics[scale=0.405]{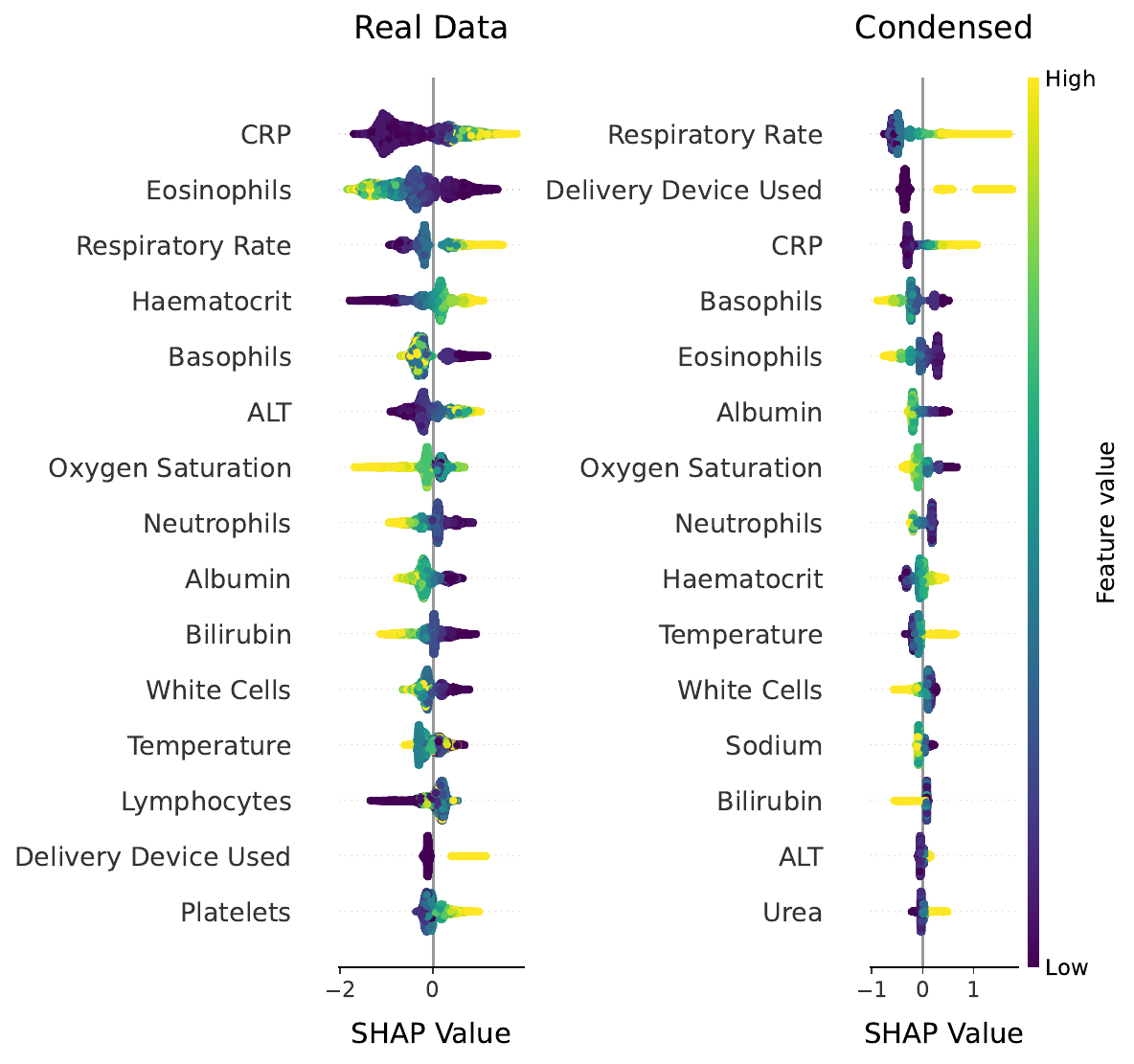}
    \caption{OUH dataset}
    \label{fig:shap_OUH}
\end{subfigure}

\vspace{0.4cm}

\begin{subfigure}[b]{0.47\textwidth}
    \centering
    \includegraphics[scale=0.405]{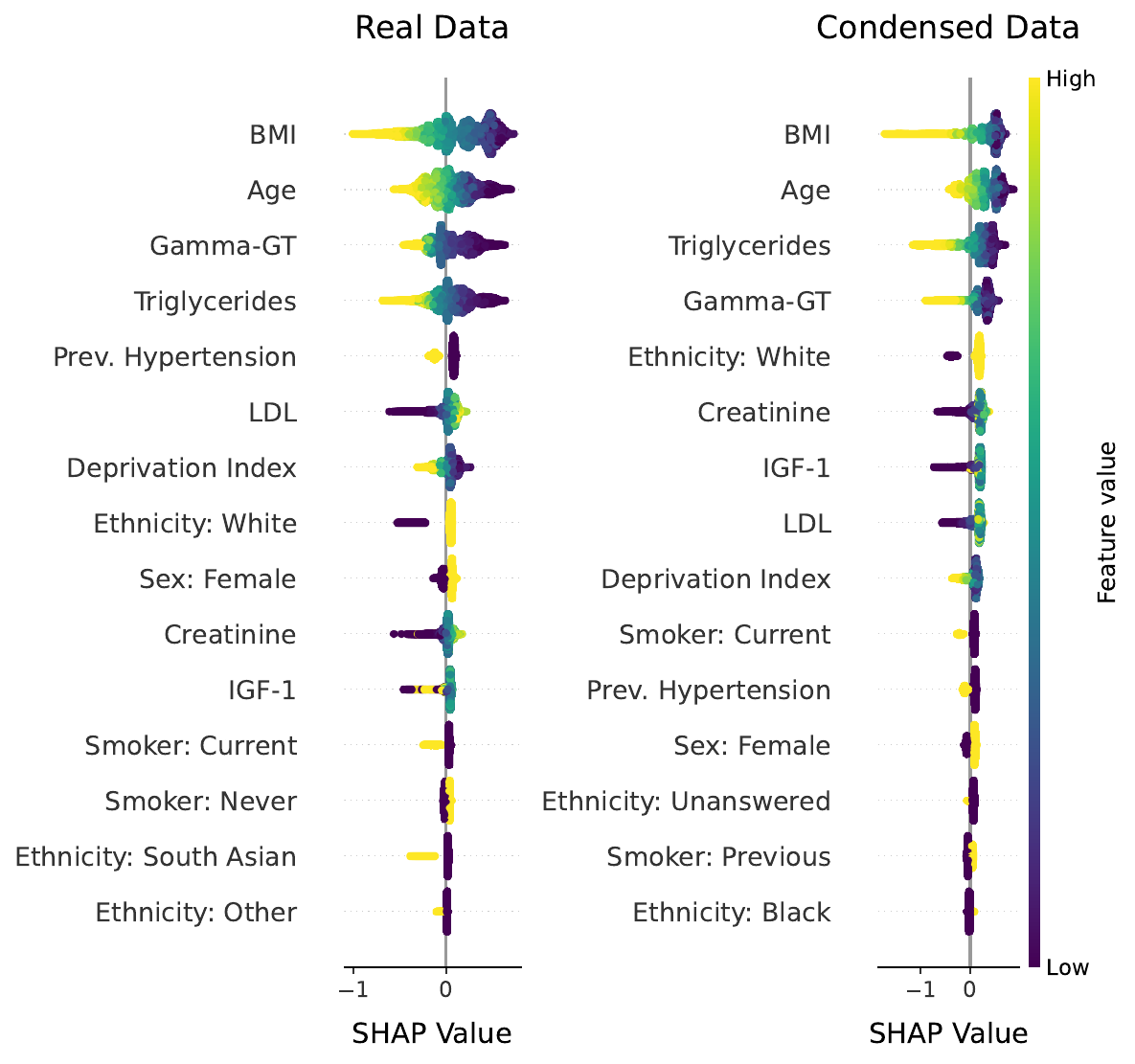}
    \caption{Diabetes}
    \label{fig:shap_diab}
\end{subfigure}
\hspace{0.02\textwidth}
\begin{subfigure}[b]{0.47\textwidth}
    \centering
    \includegraphics[scale=0.405]{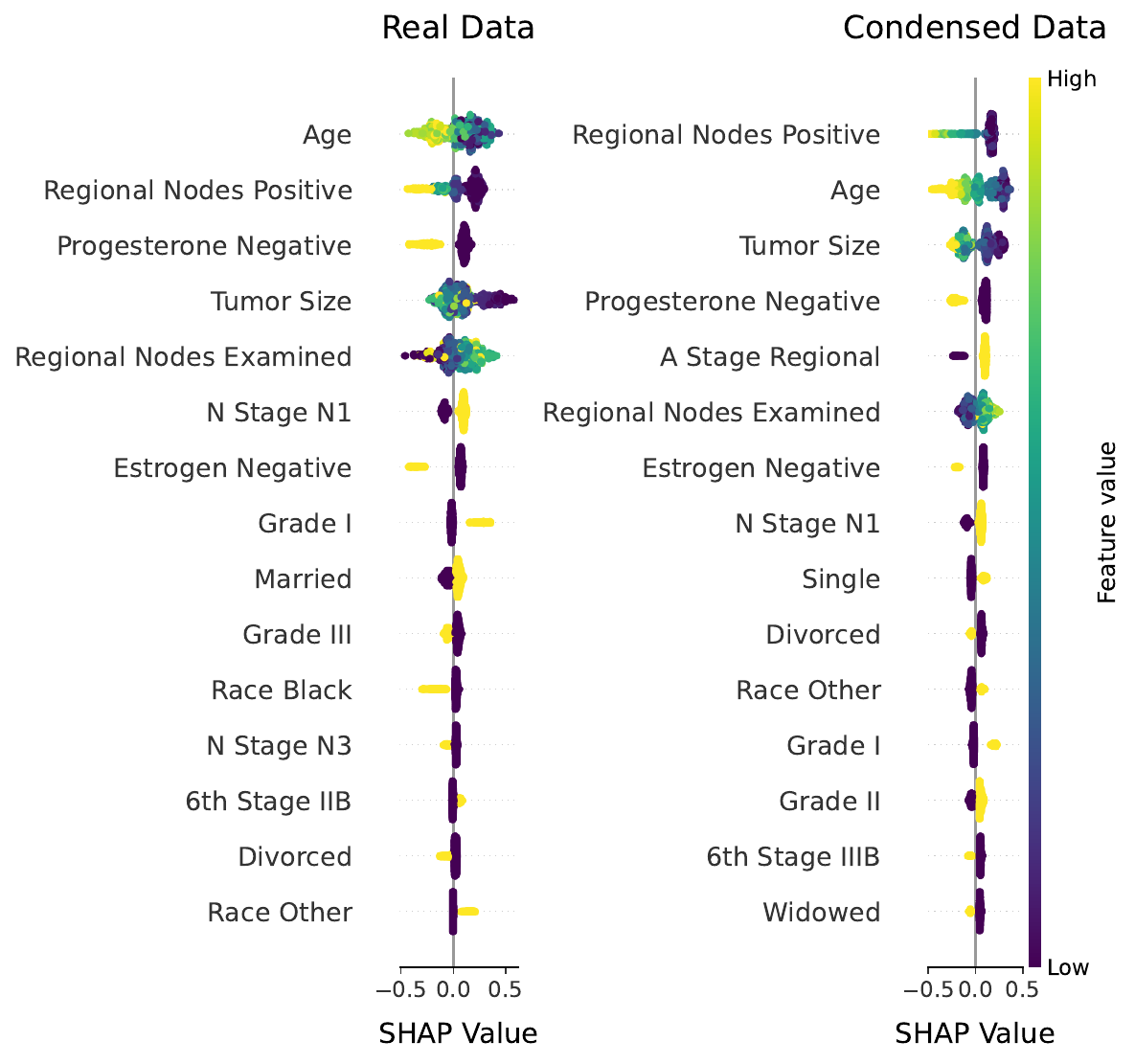}
    \caption{SEER}
    \label{fig:shap_seer}
\end{subfigure}

\caption{SHAP-based feature attribution comparison between models trained on real and condensed datasets. Top row: feature attributions from XGBoost models trained for COVID-19 prediction on (a) PUH and (b) OUH datasets. Bottom row: feature attributions from XGBoost-ARF models trained for survival prediction on (c) the UK Biobank diabetes dataset and (d) the SEER dataset. Each panel shows results using the best-performing condensed dataset.}
\label{fig:shap_combined}
\end{figure}

While we have established the predictive utility of the condensed datasets, clinical relevance also requires the feature attribution behaviour of models trained on condensed data to align closely with those trained on real data. To analyse this, we selected the best performing condensed datasets and performed Shapely additive explanations (SHAP) \cite{NIPS2017_8a20a862} based analysis on XGBoost and XGBoost-ARF models trained on the real and condensed datasets for prediction and survival analysis tasks. Figures \ref{fig:shap_combined} illustrates the SHAP-based feature attribution comparison between models trained on the real and condensed datasets.

\vspace{0.1cm}
\noindent In the COVID-19 prediction tasks using the PUH and OUH datasets, we observed strong alignment in the top SHAP-ranked features. Inflammatory markers such as C-reactive protein (CRP), eosinophils, and white blood cells, alongside vital signs like respiratory rate and temperature, were consistently highlighted by both real and condensed models. Minor differences in feature ordering, such as the elevated importance of respiratory rate and delivery device use in the condensed models, likely reflect the condensation process favouring broadly predictive and site-invariant features that generalise well across patients. In contrast, features like haemoglobin and platelets, which were more prominent in the real-data models, were down-weighted or absent in the condensed versions. These patterns suggest that condensation substitutes correlated proxies---such as haematocrit for haemoglobin or urea for renal function---to preserve clinical relevance. The SHAP analysis on the proteomics task (Figure S1 of the supplementary document) also exhibited similar behaviour, where both models highlighted overlapping proteins such as CNTN5, PAMR1, and SMNDC1, indicating preservation of core molecular signals. Minor differences, such as PTN in the real model and IGLC2 in the condensed, likely reflect redundancy among predictive markers.

\vspace{0.1cm}
\noindent A similar trend was observed in the survival analysis tasks using the Diabetes (UK Biobank) and SEER datasets. Core survival predictors---such as body mass index (BMI), age, triglycerides, and low-density lipoprotein (LDL) in predicting diabetes, and tumour size, lymph node involvement, hormone receptor status, and age in SEER---were consistently identified by both real and condensed models. In some cases, the condensed models reweighted or introduced alternative variables, such as creatinine in diabetes or widowed status in SEER, likely reflecting underlying correlations that allowed the models to approximate the same risk structure with different inputs.

\vspace{0.1cm}
\noindent In both the Diabetes and SEER survival tasks, we further compared the feature importance profiles of Cox models trained on real and condensed datasets by examining hazard ratios (HRs) documented in Table \ref{tab:hr}. The rankings of covariates were broadly consistent between real and synthetic models, with top predictors such as body mass index (BMI), triglycerides, age, and gamma-glutamyl transferase (Gamma-GT, a liver enzyme) in the diabetes dataset, and Regional Nodes Positive (number of lymph nodes with metastasis), Grade III (poorly differentiated tumour grade), age, and hormone receptor status (oestrogen/progesterone positive or negative) in the SEER task appearing across both versions. While HR magnitudes were generally attenuated in the condensed models, the direction and relative ordering of covariate effects remained aligned. 

\vspace{0.1cm}
\noindent Together, these results support the conclusion that models trained on condensed datasets exhibit feature attribution behaviour that is closely aligned with models trained on real data. The condensation process retains key clinical signals, substitutes them with correlated proxies where necessary, and avoids reliance on spurious or site-specific features preserving interpretability and trustworthiness.

\section{Discussion}
This study establishes that DC can be successfully applied to non-differentiable classical clinical models such as Cox regression and decision trees, marking a meaningful extension beyond prior work focused primarily on deep learning. Despite being substantially smaller and trained under privacy constraints, the condensed datasets retain most of the predictive performance and interpretability of their full-data counterparts, with only modest trade-offs. Across six clinical datasets, models trained on condensed data achieved near-parity with those trained on real data, suggesting that the condensation process selectively preserves patterns most relevant for downstream learning.


\vspace{0.1cm}
\noindent Beyond predictive performance, models trained on condensed data also exhibited interpretability patterns closely aligned with those learned from real data. This consistency was observed across tasks, with each model recovering key features known to be clinically relevant for its specific prediction or survival objective---as reflected in SHAP values and hazard ratios. When differences did arise, models trained on condensed data typically relied on correlated or clinically plausible alternatives, suggesting that the synthetic data preserved stable and transferable signals. Such consistency in feature attribution is essential for clinical applications, where interpretability underpins trust, regulatory acceptance, and practical deployment. By retaining both predictive accuracy and coherent attribution profiles, the condensed datasets support the reconstruction of transparent decision pathways, reinforcing their suitability for safety-critical settings.

\begin{figure}[!t]
\centering
\includegraphics[scale=0.95]{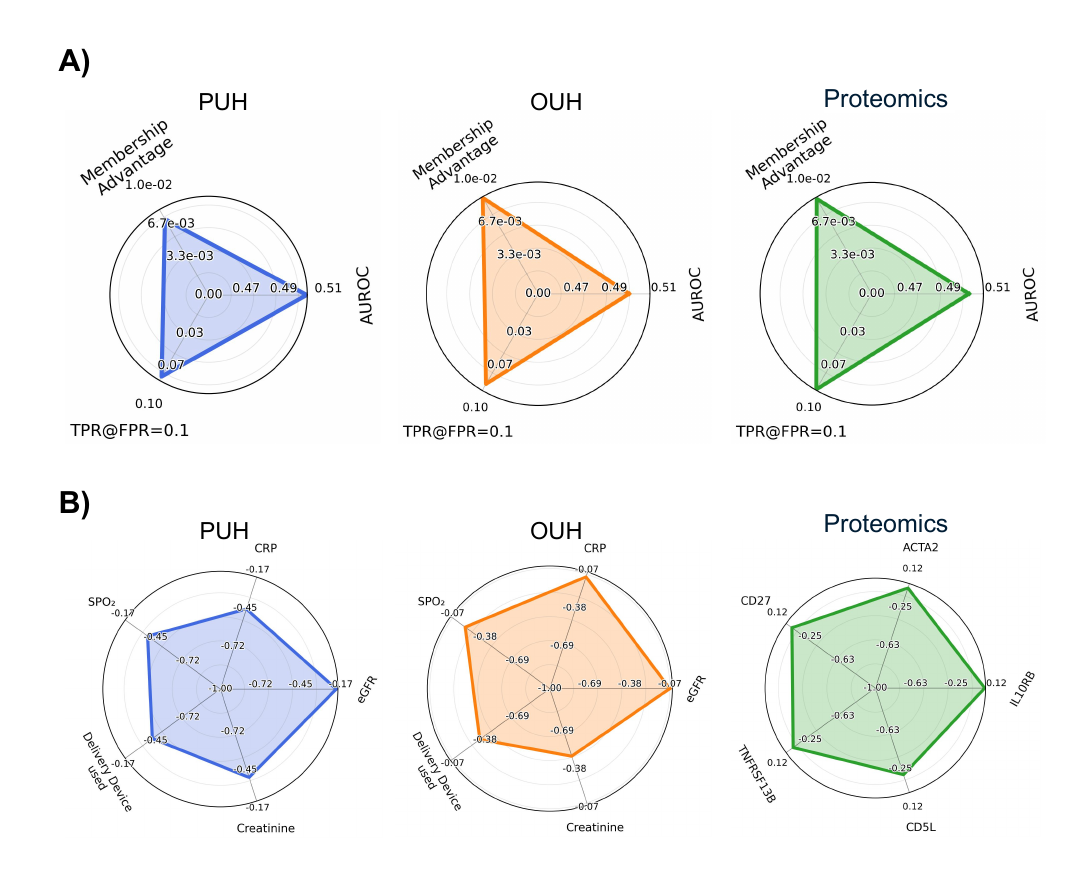}
\caption{Membership and attribute inference attacks on condensed data. \textbf{A)} Performance of a distance-based white-box membership inference attack, evaluated using AUROC, membership advantage, and true positive rate at a false positive rate of 0.1. \textbf{B)} Performance of the attribute inference attack, reported as $R^2$ scores for top attributes.}
\label{fig:inf}
\end{figure}

\vspace{0.1cm}
\noindent In addition to utility and interpretability, the framework provides robust privacy protection. Differential privacy is enforced during condensation, offering formal bounds on the influence of individual training records. To complement these guarantees, we evaluated the privacy leakage of the condensed datasets using white-box membership \cite{shokri2017membership,liu2022membership} and attribute inference attacks \cite{fredrikson2015model}. In our hypothetical scenario, only the condensed data is released, but the attacker is assumed to have full access to the real dataset, enabling a stringent evaluation. This worst-case setup tests whether condensed data alone can reveal information about individuals in the private real dataset. As shown in Figure~\ref{fig:inf}, attack performance remained near chance across all datasets: membership inference yielded negligible gains, while attribute inference produced uniformly low $R^2$ scores. These results suggest that the condensed data neither exposes individual membership nor enables reconstruction of sensitive attributes. Together with formal guarantees, this empirical resilience supports the concept of safe release of condensed datasets for downstream research and benchmarking. More broadly, it lays the groundwork for clinical data democratisation---enabling equitable access to representative, high-utility datasets, even in settings constrained by legal, infrastructural, or governance barriers to data sharing.

\vspace{0.1cm}
\noindent Traditional privacy criteria such as $\ell$-diversity \cite{machanavajjhala2007diversity} and $t$-closeness \cite{li2006t} are not evaluated in this study, as they are designed for structured tabular datasets that retain explicit quasi-identifiers (e.g., age, ZIP code) and rely on partitioning records into equivalence classes. In contrast, the condensed data is learned from pseudonymised clinical datasets in which such identifiers have largely been removed, and the resulting synthetic examples are not direct transformations of individual records. As a result, these criteria are not directly applicable. Instead, our framework relies on differential privacy to provide formal, instance-level protection against inference attacks, even in the presence of arbitrary auxiliary information.

\vspace{0.1cm}
\noindent As mentioned earlier, DC differs fundamentally from other privacy-preserving paradigms such as federated learning (FL) and generative modelling. In the context of clinical ML, these differences translate into tangible advantages for scalable, auditable, and equitable data sharing. While FL enables decentralised training, it depends on tightly coordinated infrastructure and shared governance---practical barriers that often limit broader participation. More critically, it produces no reusable data artifact: neither external researchers nor downstream developers can inspect, reuse, or adapt the underlying data in any form. In contrast, dataset condensation produces a static, self-contained surrogate that can be reused, shared, and audited offline which enables equitable access to high-utility clinical data without compromising individual privacy.


\vspace{0.1cm}
\noindent Generative models, though capable of producing high-fidelity samples, face intrinsic trade-offs between realism and task-specific utility. By design, models such as GANs and diffusion-based generators aim to replicate the full data distribution, which can amplify irrelevant variation and underrepresent rare but clinically important cases such as subtle haemorrhages in diabetic retinopathy \cite{zhou2020dr} or minority subgroups affected by distributional shifts in radiology datasets \cite{ktena2024generative}. This misalignment arises because generative objectives typically prioritise distributional fidelity over decision-relevant structure. In practice, these models also require large training datasets, extensive hyperparameter tuning, and careful regularisation to mitigate overfitting. Even under differential privacy, they may remain vulnerable to memorisation, particularly when trained with weaker privacy guarantees (e.g., high $\varepsilon$) or limited data \cite{chen2020gan}. Their objective to approximate the full data manifold, combined with the ability to generate an unbounded number of samples, increases the risk of reproducing not only exact records but also semantically similar or rare patterns potentially compromising privacy even when direct duplication is avoided \cite{hayes2019logan}. In contrast, dataset condensation directly optimises a fixed, task-aligned subset of synthetic examples for predictive performance under formal privacy constraints. This focused objective reduces sample complexity, limits the risk of memorisation, and better supports clinically meaningful generalisation.


\vspace{0.1cm}
\noindent While this study demonstrates the feasibility of differentially private DC for classical clinical models, some limitations define the scope of its current applicability. First, although the condensed datasets generalised well across clinical sites and downstream models, performance was strongest when the downstream models shared inductive biases with the pre-trained model used during condensation, suggesting an opportunity to improve model-agnosticity. This limitation could potentially be addressed by incorporating multiple pre-trained models into the condensation process to encourage broader downstream compatibility. Second, although our empirical evaluations used strong white-box attacks and revealed minimal leakage, future work should assess robustness to more adversarial conditions such as adaptive attacks, repeated querying, and integration with auxiliary information or external systems. Finally, while the proposed framework is model-agnostic and readily compatible with neural networks, we focused exclusively on classical models such as decision trees and Cox regression, which are widely used in healthcare due to their interpretability and established clinical relevance. Extending the method to deep models is technically straightforward and represents a natural next step for future work.

\vspace{0.1cm}
\noindent Despite these limitations, this work demonstrates that dataset condensation offers a scalable and practical solution for generating compact, task-effective datasets across a wide range of predictive tasks in healthcare. By enabling the release of synthetic data that preserves downstream utility while enforcing formal privacy guarantees, the proposed framework supports more equitable and reproducible clinical machine learning. Beyond methodological advances, this approach has tangible applications in real-world settings such as enabling model development in data-restricted environments, facilitating external benchmarking across institutions, and supporting regulatory evaluation through privacy-preserving surrogates of sensitive datasets. In particular, it points to a promising direction for global data democratisation: condensed datasets derived from high-resource health systems could be safely shared with institutions in low- and middle-income countries, helping to lower structural barriers to participation in clinical AI research. While not a definitive solution, this work lays important groundwork for enabling more inclusive, transparent, and privacy-preserving approaches to clinical machine learning.

\section{Method}



\subsection{Model Training}
To supervise data condensation, we trained predictive models on the real training dataset for both binary classification and survival analysis tasks. For binary classification, we used a gradient-boosted decision tree model (XGBoost) trained to minimise logistic loss. For survival analysis, we employed two models: a Cox proportional hazards model, and an accelerated failure time (AFT) model implemented using XGBoost. The AFT model assumes that the logarithm of the event time follows a normal distribution, corresponding to a parametric log-normal survival model. All survival times were scaled using robust normalization:

\begin{equation}
\tilde{t} = \frac{t}{s}, \quad \text{where} \quad
s = 
\begin{cases}
\text{IQR}(t_{\text{event}=1}) & \text{if } \text{IQR}(t_{\text{event}=1}) > 0 \\
\text{Median}(t_{\text{event}=1}) & \text{otherwise}
\end{cases}
\end{equation}
Here, $\tilde{t}$ denotes the scaled time-to-event, and $t_{\text{event}=1}$ refers to durations associated with uncensored events. This scaling improves numerical stability in survival loss optimization, particularly in the AFT objective.

\noindent All model hyperparameters were tuned using validation set performance, based on AUROC for classification and C-index for survival analysis.

\subsection{Dataset Condensation via Zero-order Gradient Estimation}
Given a clinical dataset \(\mathcal{X}_{\text{real}} = (\mathbf{X}, \mathbf{y})\), where \(\mathbf{X} \in \mathbb{R}^{n \times d}\) denotes a matrix of \(n\) examples each with \(d\) features, and \(\mathbf{y} \in \mathbb{R}^{n}\) is the corresponding vector of target labels or outcomes. The goal of dataset condensation is to learn a smaller synthetic dataset \(\mathcal{X}_{\text{syn}} = (\mathbf{X}_{\text{syn}}, \mathbf{y}_{\text{syn}})\), where \(\mathbf{X}_{\text{syn}} \in \mathbb{R}^{m \times d}\) with \(m \ll n\), and \(\mathbf{y}_{\text{syn}} \in \mathbb{R}^{m}\) contains the associated synthetic labels.

\vspace{0.1cm}
\noindent To achieve this, we first train a model \(f_{\text{real}}(\cdot)\)---XGBoost in our case---on the real dataset \(\mathcal{X}_{\text{real}}\). The synthetic inputs \(\mathbf{X}_{\text{syn}}\) are then randomly initialised, and the corresponding synthetic labels \(\mathbf{y}_{\text{syn}}\) are constructed to reflect the target distribution of the task. For binary classification, \(\mathbf{y}_{\text{syn}}\) is initialised with an equal number of 0s and 1s to ensure class balance, or according to a desired ratio to model class imbalance. Finally, we optimise or train \(\mathbf{X}_{\text{syn}}\) so that a model trained on the synthetic dataset exhibits similar predictive behaviour to a model trained on the real dataset, as measured on an external validation or test set.  

\vspace{0.1cm}
\noindent To train \(\mathbf{X}_{\text{syn}}\), we define a composite loss function comprising a binary cross-entropy (BCE) loss and a class-conditional prediction distribution matching loss, both evaluated using a fixed pretrained model \(f_{\text{real}}(\cdot)\). The binary cross-entropy term encourages the synthetic inputs \(\mathbf{X}_{\text{syn}}\) to produce predictions consistent with their assigned labels \(\mathbf{y}_{\text{syn}}\), while the distribution matching term aligns the average model outputs on synthetic data with those on real data within each class, thereby preserving the class-conditional output behaviour of the original dataset.

\vspace{0.1cm}
\noindent The BCE loss is defined as: 

\begin{equation}
    \ell_{\text{pred}}(\mathbf{X}_{\text{syn}}, \mathbf{y}_{\text{syn}}) = \frac{1}{m} \sum_{i=1}^{m} \left[ -y_i \log f_{\text{real}}(\mathbf{x}_i) - (1 - y_i) \log \left(1 - f_{\text{real}}(\mathbf{x}_i) \right) \right],
\end{equation}
where \(\mathbf{x}_i \in \mathbf{X}_{\text{syn}}\) represents $i$th row or example and \(y_i \in \mathbf{y}_{\text{syn}}\) represents the corresponding label. \noindent Similarly, the class-conditional prediction distribution matching loss \(\ell_{\text{match}}\) encourages the pretrained model’s average predictions on synthetic data to align with those on real data within each class:
\begin{equation}
\ell_{\text{match}}(\mathbf{X}_{\text{syn}}, \mathbf{y}_{\text{syn}}, \mathbf{X}_{\text{real}}, \mathbf{y}_{\text{real}}) = \sum_{c \in \{0,1\}} \left| \frac{1}{|\mathcal{I}^{\text{syn}}_c|} \sum_{i \in \mathcal{I}^{\text{syn}}_c} f_{\text{real}}(\mathbf{x}^{\text{syn}}_i) - \frac{1}{|\mathcal{I}^{\text{real}}_c|} \sum_{j \in \mathcal{I}^{\text{real}}_c} f_{\text{real}}(\mathbf{x}^{\text{real}}_j) \right|,
\end{equation}

\noindent where \(\mathbf{x}^{\text{syn}}_i \in \mathbf{X}_{\text{syn}}\) and \(\mathbf{x}^{\text{real}}_j \in \mathbf{X}_{\text{real}}\) are samples from the synthetic and real datasets, respectively. The index sets are defined as \(\mathcal{I}^{\text{syn}}_c = \{i : y^{\text{syn}}_i = c\}\) and \(\mathcal{I}^{\text{real}}_c = \{j : y^{\text{real}}_j = c\}\). Then, the composite loss function is: 
\begin{equation}
    \ell = \ell_{\text{pred}}(\mathbf{X}_{\text{syn}}, \mathbf{y}_{\text{syn}}) + \alpha .\, \ell_{\text{match}}(\mathbf{X}_{\text{syn}}, \mathbf{y}_{\text{syn}}, \mathbf{X}_{\text{real}},\mathbf{y}_{\text{real}}).
    \label{eq:loss}
\end{equation}
Here $\alpha$ is an adaptive weighting factor defined as:
\begin{equation}
\alpha = \left( \frac{\ell_{\text{pred}}}{\ell_{\text{match}} + \epsilon} \right) \cdot \left( \frac{\rho}{1 - \rho} \right),
\end{equation}
where $\rho$ is a target loss ratio (e.g., 0.1) and $\epsilon$ is a small constant for numerical stability. This formulation allows $\alpha$ to dynamically balance the contribution of the two losses, ensuring that the matching loss contributes approximately a fraction $\rho$ of the total loss throughout training. This is particularly helpful when the magnitudes of the prediction and matching losses differ substantially, preventing one from dominating the optimization.

\vspace{0.1cm}
\noindent To optimise the synthetic dataset \(\mathbf{X}_{\text{syn}} \in \mathbb{R}^{m \times d}\), we require the gradient of the composite loss \(\ell\) with respect to the synthetic inputs. Since this loss depends on the model’s predictions, we apply the chain rule of differentiation to decompose the required gradient into two components: the sensitivity of the loss to the model’s outputs, and the sensitivity of the model’s outputs to the inputs. Formally, the required gradient $\nabla_{\mathbf{X}_{\text{syn}}} \ell$ is given by:
\begin{equation}
\nabla_{\mathbf{X}_{\text{syn}}} \ell =
\left( \frac{\partial \ell}{\partial f_{\text{real}}(\mathbf{X}_{\text{syn}})} \right)
\cdot
\left( \frac{\partial f_{\text{real}}(\mathbf{X}_{\text{syn}})}{\partial \mathbf{X}_{\text{syn}}} \right).
\label{eq:grad}
\end{equation}
Here, the first term is analytically computable because the loss function \(\ell\) is differentiable with respect to the model outputs. Since we have explicit access to both the predicted values \(f_{\text{real}}(\mathbf{X}_{\text{syn}})\) and their corresponding labels \(\mathbf{y}_{\text{syn}}\), this gradient can be evaluated directly without requiring gradients through the model itself.
However, because the model \(f_{\text{real}}(\cdot)\) is non-differentiable, the second term cannot be obtained via backpropagation. We therefore approximate it using symmetric finite differences, enabling end-to-end optimisation of the synthetic data despite the model’s non-differentiability. However, because the model \(f_{\text{real}}(\cdot)\) is non-differentiable, the second term cannot be computed using automatic differentiation (i.e. backpropagation), which is the standard technique used in neural networks to compute gradients. Instead, we approximate this term using symmetric finite differences, allowing the synthetic data to be optimised end-to-end despite the model's non-differentiability.

\vspace{0.1cm}
\noindent To estimate $\frac{\partial f_{\text{real}}(\mathbf{X}_{\text{syn}})}{\partial \mathbf{X}_{\text{syn}}}$, we apply symmetric finite differences to each feature dimension of the synthetic batch. For a small perturbation \(\epsilon_j\) and a unit vector \(\mathbf{e}_j\) in the \(j\)-th input dimension, the partial derivative with respect to feature \(j\) (i.e., the \(j\)-th column of \(\mathbf{X}_{\text{syn}}\), representing that feature across all synthetic examples) is approximated as:
\begin{equation}
\frac{\partial f_{\text{real}}(\mathbf{X}_{\text{syn}})}{\partial \mathbf{X}_{\text{syn}, j}} \approx \frac{f_{\text{real}}(\mathbf{X}_{\text{syn}} + \epsilon_j \mathbf{E}_j) - f_{\text{real}}(\mathbf{X}_{\text{syn}} - \epsilon_j \mathbf{E}_j)}{2\epsilon_j}.
\label{eq:zero}
\end{equation}
Here, \(\mathbf{E}_j\) is a perturbation matrix with the \(j\)-th column set to \(\mathbf{e}_j\) and all other entries zero, so that only the \(j\)-th feature of all examples is perturbed while others remain unchanged. This procedure is repeated for all feature dimensions \(j = 1, \dots, d\), in order to construct the full Jacobian matrix \(\frac{\partial f_{\text{real}}(\mathbf{X}_{\text{syn}})}{\partial \mathbf{X}_{\text{syn}}} \), which forms the second term in the gradient expression $\nabla_{\mathbf{X}_{\text{syn}}} \ell$ as defined in Equation \ref{eq:grad}. These zero-order gradient estimates \(\nabla_{\mathbf{X}_{\text{syn}}}\ell\) are then passed to a first-order optimiser (Adam in our experiments) to refine the synthetic data.  Concretely, at iteration \(t\) we perform:
\begin{equation}
\mathbf{X}_{\text{syn}}^{(t+1)}
=\mathbf{X}_{\text{syn}}^{(t)}
-\eta\,\nabla_{\mathbf{X}_{\text{syn}}}\ell,
\label{eq:gd}
\end{equation}
where \(\eta_t\) is the learning rate or step size.

\subsection{Extension to Survival Analysis}

The framework extends naturally from classification to survival analysis, where the objective involves predicting time-to-event outcomes under censoring. Given a real survival dataset \(\mathcal{X}_{\text{real}} = (\mathbf{X}, \mathbf{T}, \mathbf{E})\), where \(\mathbf{X} \in \mathbb{R}^{n \times d}\) represents \(n\) examples with \(d\) features, \(\mathbf{T} \in \mathbb{R}^n\) denotes time-to-event values, and \(\mathbf{E} \in \{0, 1\}^n\) indicates whether an event was observed (\(1\)) or censored (\(0\)), the goal is to generate a condensed synthetic dataset \(\mathcal{X}_{\text{syn}} = (\mathbf{X}_{\text{syn}}, \mathbf{T}_{\text{syn}}, \mathbf{E}_{\text{syn}})\), where \(\mathbf{X}_{\text{syn}} \in \mathbb{R}^{m \times d}\) and \(m \ll n\).

\vspace{0.2cm}
\noindent In contrast to classification, survival outcomes require initialising both time-to-event values \(\mathbf{T}_{\text{syn}}\) and censoring indicators \(\mathbf{E}_{\text{syn}}\). To ensure sufficient signal and variability during optimisation, we initialise half of the synthetic samples as uncensored (i.e., with \(E_i = 1\) for entries in \(\mathbf{E}_{\text{syn}}\)), assigning event times \(\mathbf{T}_{\text{syn}}\) drawn uniformly from stratified bins spanning the range between the minimum observed event time and the maximum censoring time in the real dataset. The remaining half are assigned as censored (i.e., \(E_i = 0\)), with their times fixed to the maximum observed censoring time. This strategy ensures a balanced representation of censored and uncensored cases, while maintaining a structured temporal signal that supports stable training.

\vspace{0.2cm} 
\noindent To train the condensed dataset for survival analysis, we employ a composite loss comprising a model-specific supervision loss and a distribution matching term that aligns prediction distributions between real and synthetic data. The form of \(\ell_{\text{supervision}}\) depends on the type of survival model used. For Cox models, which output relative risk scores under the proportional hazards assumption, we supervise the synthetic dataset using the negative partial log-likelihood, evaluated over uncensored synthetic samples:
\begin{equation}
\ell_{\text{cox}} = -\sum_{i: E_i = 1} \left( f_{\text{real}}(\mathbf{x}_i) - \log \sum_{j \in \mathcal{R}(T_i)} \exp(f_{\text{real}}(\mathbf{x}_j)) \right),
\end{equation}
where \( \mathbf{x}_i \in \mathbf{X}_{\text{syn}} \), \( T_i \in \mathbf{T}_{\text{syn}} \), and \( E_i \in \mathbf{E}_{\text{syn}} \). The risk set \(\mathcal{R}(T_i)\) consists of all synthetic examples \(j\) with \(T_j \geq T_i\), i.e., those still at risk at time \(T_i\).  The term inside the summation compares the predicted log-risk for sample \(i\) to the aggregated risk scores of all at-risk synthetic examples. This structure encourages the model to assign higher predicted risks to samples with earlier events, thereby preserving the rank-based ordering of event times consistent with the Cox proportional hazards framework.

\noindent For AFT models, which predict log-survival times, we supervise the synthetic dataset using a smoothed regression objective that compares the model's predictions to the log-transformed synthetic survival times:
\begin{equation}
\ell_{\text{aft}} = \frac{1}{m} \sum_{i=1}^{m} \text{SmoothL1}\left( f_{\text{real}}(\mathbf{x}_i), \log T_i \right),
\end{equation}
where \( \mathbf{x}_i \in \mathbf{X}_{\text{syn}} \) and \( T_i \in \mathbf{T}_{\text{syn}} \) denote the \(i\)-th synthetic feature vector and its assigned time-to-event label, respectively. The Smooth L1 loss, also known as the Huber loss provides a numerically stable surrogate for the intractable log-likelihood of parametric survival distributions, particularly in settings where gradients are estimated via finite differences.

\noindent Although the supervision loss differs between Cox and AFT models, the distribution matching loss \(\ell_{\text{match}}\) is identical. Unlike in classification, where matching is performed per class, survival models require stratifying predictions by survival characteristics. Specifically, we partition the synthetic predictions into \(K\) quantile-based strata based on \(\mathbf{T}_{\text{syn}}\) and \(\mathbf{E}_{\text{syn}}\), and align the mean predictions within each stratum to those of the corresponding real strata:
\begin{equation}
\ell_{\text{match}} = \frac{1}{K} \sum_{k=1}^{K} \left| \frac{1}{|\mathcal{I}^{\text{syn}}_k|} \sum_{i \in \mathcal{I}^{\text{syn}}_k} f_{\text{real}}(\mathbf{x}^{\text{syn}}_i) 
- 
\frac{1}{|\mathcal{I}^{\text{real}}_k|} \sum_{j \in \mathcal{I}^{\text{real}}_k} f_{\text{real}}(\mathbf{x}^{\text{real}}_j) \right|
\end{equation}

\noindent where \(\mathbf{x}^{\text{syn}}_i \in \mathbf{X}_{\text{syn}}\) and \(\mathbf{x}^{\text{real}}_j \in \mathbf{X}_{\text{real}}\) are synthetic and real input samples, respectively, and \(\mathcal{I}^{\text{syn}}_k\) and \(\mathcal{I}^{\text{real}}_k\) denote the sets of indices assigned to stratum \(k\). This stratified matching ensures that the synthetic dataset captures key patterns in the distribution of predicted risks or survival times, preserving population-level survival structure across heterogeneous risk groups.

\noindent These two respective losses are combined as explained in Equation \ref{eq:loss} to compute the final composite loss. 

\vspace{0.1cm}
\noindent Finally, the process of computing gradients and training $\mathbf{X}_{\text{syn}}$ is identical to the one explained for the classification prediction scenario.

\subsection{Differential Privacy Mechanism}

We ensure formal \((\varepsilon, \delta)\)-differential privacy during dataset condensation by perturbing the estimated gradient matrix \(\nabla_{\mathbf{X}_{\text{syn}}} \ell \in \mathbb{R}^{m \times d}\), computed via finite differences due to the non-differentiable nature of the real model \(f_{\text{real}}(\cdot)\). For each synthetic example \(\mathbf{x}_i \in \mathbb{R}^d\), let \(\mathbf{g}_i \approx \nabla_{\mathbf{x}_i} \ell\) denote the corresponding estimated gradient. To bound sensitivity, each gradient is \(\ell_2\)-clipped to a predefined norm \(C\):
\begin{equation}
\mathbf{g}_i^{\mathrm{clip}} = \mathbf{g}_i \cdot \min\left(1, \frac{C}{\|\mathbf{g}_i\|_2} \right).
\end{equation}
To preserve utility while ensuring privacy, Gaussian noise is added adaptively. Specifically, the noise standard deviation \(\sigma_i\) is selected from a discrete set \(\{0.25\sigma_{\text{base}}, 0.5\sigma_{\text{base}}, \ldots, 2\sigma_{\text{base}}\}\), such that the signal-to-noise ratio (SNR) remains at least 1 for effective utility, and \(\sigma_i \geq 0.25\sigma_{\text{base}}\) to guarantee a minimum privacy level:
\begin{equation}
\tilde{\mathbf{g}}_i = \mathbf{g}_i^{\mathrm{clip}} + \mathcal{N}(0, \sigma_i^2 C^2 \mathbf{I}).
\end{equation}
The complete differentially private gradient matrix is then obtained by stacking the noised per-example gradients:
\begin{equation}
\tilde{\nabla}_{\mathbf{X}_{\text{syn}}} \ell =
\begin{bmatrix}
\tilde{\mathbf{g}}_1^\top \\
\tilde{\mathbf{g}}_2^\top \\
\vdots \\
\tilde{\mathbf{g}}_m^\top
\end{bmatrix}.
\end{equation}
This $\tilde{\nabla}_{\mathbf{X}_{\text{syn}}} \ell$ is the differentially-private variant of $\nabla_{\mathbf{X}_{\text{syn}}} \ell$ and is used to update \(\mathbf{X}_{\text{syn}}\) as defined in Equation \ref{eq:gd}.\\

\noindent \textbf{Privacy accounting:} We track cumulative privacy loss using Rényi Differential Privacy (RDP), assuming Poisson subsampling of real data at each optimisation step. Specifically, \(m\) real data points are sampled independently with replacement from a total of \(N_{\text{real}}\) records, resulting in a sampling rate \(q = m/N_{\text{real}}\). The RDP accountant is invoked once per step using the actual noise multiplier \(\tilde{\sigma}_{\text{step}} = \sigma_{\text{step}} / C\) and sampling rate \(q\). This approach yields tight composition without relying on worst-case approximations. The total privacy cost is then converted to an \((\varepsilon, \delta)\)-DP guarantee using standard RDP-to-DP conversion, with \(\delta\) fixed at \(10^{-5}\). Accounting is performed up to the best early-stopped iteration, ensuring a valid and conservative estimate of the final privacy budget.

\subsection{White-box Membership Inference Attack}
To assess potential privacy leakage introduced by dataset condensation, we implement a white-box membership inference attack (MIA) under the assumption that the adversary has full access to the released synthetic dataset. The objective is to determine whether a given real data point \(\mathbf{x} \in \mathbf{X}_{\text{real}}\) was included in the private training set used to generate the synthetic dataset (i.e., a member), or was held out (i.e., a non-member). This attack leverages the hypothesis that training examples may lie closer to the synthetic data manifold than unseen test samples, thereby revealing potential memorisation in the condensation process.

\vspace{0.1cm}
\noindent To implement this attack, we use a distance-based strategy that relies on the intuition that training samples are more tightly embedded within the synthetic data manifold than unseen points. Specifically, for each real data point \(\mathbf{x} \in \mathbf{X}_{\text{real}}\), we compute its distances to the \(k\) nearest neighbours within the synthetic dataset \(\mathbf{X}_{\text{syn}}\), using three distance metrics: Euclidean, Manhattan, and Cosine. For each metric, we extract five summary statistics from the resulting distance vector: the mean, minimum, maximum, standard deviation, and range. This yields a 15-dimensional feature vector per sample (5 statistics \(\times\) 3 metrics). As an illustrative example, the average Euclidean distance is computed as:
\begin{equation}
d_k(\mathbf{x}) = \frac{1}{k} \sum_{j=1}^{k} \left\lVert \mathbf{x} - \operatorname{NN}_j(\mathbf{x}; \mathbf{X}_{\text{syn}}) \right\rVert_2,
\end{equation}
where \(\operatorname{NN}_j(\mathbf{x}; \mathbf{X}_{\text{syn}})\) denotes the \(j\)th nearest neighbour of \(\mathbf{x}\) in the synthetic dataset. We set \(k = 5\) in all experiments, although alternative values yielded qualitatively similar results. These refined features serve as input to a binary classifier trained to distinguish members (\(\mathbf{x} \in \mathbf{X}_{\text{real}}^{\text{train}}\)) from non-members (\(\mathbf{x} \in \mathbf{X}_{\text{real}}^{\text{test}}\)). Under the framework of \((\varepsilon, \delta)\)-differential privacy, the membership advantage of any adversary is theoretically bounded by \((e^{\varepsilon} - 1)/(e^{\varepsilon} + 1) + \delta\). As such, AUROC values near 0.5 and small membership advantages indicate limited privacy leakage, while higher values may suggest memorisation of training-specific information.

\vspace{0.1cm}
\noindent The attack dataset is constructed by extracting these features for both member and non-member real samples, and labeling them accordingly: \( y = 1 \) if \( \mathbf{x} \) is a member (i.e., used during condensation) and \( y = 0 \) otherwise. We split the resulting dataset into an 80/20 stratified training and testing partition and train a Gradient Boosting classifier to distinguish members from non-members. Gradient Boosting is well-suited for this task due to its ability to capture nonlinear interactions across the multiple distance-derived features. The attacker model is trained with 100 trees, a maximum depth of 3, and a learning rate of 0.1. We repeat each attack five times with different random splits and report average performance.

\vspace{0.1cm}
\noindent  We evaluate the attacker on the held-out test partition using the following metrics:
\begin{itemize}
    \item Area under the ROC curve (AUROC), which quantifies the attacker’s discrimination ability across all thresholds;
\item Membership advantage, defined as the maximum gap between the true positive rate (TPR) and false positive rate (FPR) over all classification thresholds applied to the predicted probabilities. Formally, \(\max_{\tau} \left[ \text{TPR}(\tau) - \text{FPR}(\tau) \right]\), where \(\tau\) denotes the decision threshold.

    \item TPR at FPR = 0.1, measuring the sensitivity of the attack at a fixed 10\% false positive rate.
\end{itemize}

\subsection{Attribute Inference Attack}

In addition to membership inference, we assess the potential of dataset condensation to expose sensitive attribute information through attribute inference attacks (AIA). The objective of this attack is to determine whether a specific feature of interest (e.g., a clinical biomarker or treatment indicator) can be accurately inferred from the remaining features in the condensed dataset, thereby revealing unintended leakage of private information.

\vspace{0.1cm}
\noindent In the absence of traditional identity-linked sensitive attributes such as age or gender, we focus on contextually sensitive clinical variables that, if inferred, could disclose information about a patient’s health status or care decisions. For the OUH and PUH datasets, we target five such variables: CRP, eGFR, oxygen saturation (SPO\textsubscript{2}), creatinine, and the type of delivery device used. These features reflect inflammatory response, renal function, respiratory status, metabolic health, and treatment interventions, respectively—all of which may be considered sensitive in clinical settings. For the Proteomics dataset, we evaluate five protein markers: ACTA2, CD27, CD5L, IL10RB, and TNFRSF12B. These proteins are associated with disease pathophysiology and represent sensitive molecular signatures in biomedical research contexts.

\vspace{0.1cm}
\noindent To implement this attack, we construct a separate prediction task for each selected attribute. In each case, the target attribute is removed from the feature set, and a regression model is trained on the condensed data to predict the held-out value. The attacker is assumed to have access only to the released condensed dataset and not the real data distribution. The trained model is then evaluated on real samples to measure generalisation and potential leakage.

\vspace{0.1cm}
\noindent We consider both binary and continuous attributes in the original data; however, since all attributes are represented as continuous values in the condensed dataset, we adopt a unified regression-based approach for consistency. Specifically, we employ a gradient boosting regression model configured with 100 estimators, a learning rate of 0.1, and a maximum tree depth of 3. This design choice reflects the observation that condensation typically produces soft approximations of feature values, even for originally categorical variables. For binary attributes, the predicted continuous outputs are post-processed by thresholding at 0.5 to enable evaluation using classification metrics.

\vspace{0.1cm}
\noindent The attacker’s performance is evaluated using the coefficient of determination (R\textsuperscript{2}) for all targets. For originally binary attributes, we additionally report accuracy and area under the ROC curve (AUROC) after applying the threshold. Higher R\textsuperscript{2} or AUROC values indicate increased vulnerability to attribute inference, while values closer to zero (for R\textsuperscript{2}) or 0.5 (for AUROC) suggest effective privacy preservation. All attacks are repeated across five random seeds, and average results are reported.

\subsection{Implementation details}
Across all datasets, we generate class-balanced synthetic datasets by allocating an equal number of synthetic samples to each class. During condensation, the entire set of synthetic samples is treated as a single training batch. To compute the matching loss, an equal number of real samples are randomly drawn from the corresponding class in the real dataset during each training iteration. In composite loss, the $\rho$ is kept fixed at $0.1$ as it provided best validation scores across all datasets. The noise \(\epsilon\) used to estimate gradients in Equation~\ref{eq:zero} is sampled from a fixed range at each iteration. This range varies across datasets and is selected based on the magnitude of real-valued features and the characteristics of the target model. In most cases, we use a uniform range of \((0.025, 2)\), which was empirically found to provide stable and meaningful gradient estimates. For applying estimated gradients, we employed Adam optimiser with a fixed learning rate of $0.001$.

\vspace{0.1cm}
\noindent For all classification tasks, gradient-boosted decision trees were trained with a learning rate of 0.1, a maximum tree depth of 5, and a subsampling ratio of 0.7 to mitigate overfitting. Each model was trained for up to 100 boosting rounds and evaluated using either log-loss or area under the ROC curve, depending on the prediction task. For survival analysis, we used the same gradient-boosted tree configuration as in the classification tasks but replaced the objective with the accelerated failure time (AFT) loss. We assumed a normal distribution for log survival times and set the scale parameter of this distribution to 1.0. Cox proportional hazards models were trained with $\ell_2$-regularisation, using a regularisation strength of 1. Class weighting was applied where appropriate to address label imbalance. All models were configured based on standard practice and validated on held-out development or validation sets.

\newpage 
\section*{Data availability}
Data from OUH studied here are available from the Infections in Oxfordshire Research Database (\href{https://oxfordbrc.nihr.ac.uk/research-themes/modernising-medical-microbiology-and-big-infection-diagnostics/iord-about/}{https://oxfordbrc.nihr.ac.uk/\allowbreak research-themes/modernising-medical-microbiology-and-big-infection-diagnostics/iord-about/}), subject to an application meeting the ethical and governance requirements of the Database. Data from UHB, PUH and BH are available on reasonable request to the respective trusts, subject to HRA requirements. The SEER dataset used in this work is available at \url{https://ieee-dataport.org/open-access/seer-breast-cancer-data}. The Diabetes and Proteomics data are available at the UK Biobank. 

\section*{Code availability} The demo code is available at \url{https://github.com/AnshThakur/Zero-Order-DC}.

\section*{Acknowledgement}
DAC is supported by the Pandemic Sciences Institute at the University of Oxford; the National Institute for Health Research (NIHR)
Oxford Biomedical Research Center (BRC); an NIHR Research Professorship; a Royal Academy of Engineering Research Chair; the Well-
come Trust; the UKRI; and the InnoHK Hong Kong
Center for Center for Cerebro-cardiovascular Engineering (COCHE).

\section*{Author Contributions}
AT and DC conceived the project. AT, SM, PCN, FL, DB and LC conceived and designed the study, performed the data analysis, and prepared the manuscript. JF, LC and AS processed the UK Biobank and the CURIAL datasets. All authors contributed to results interpretation, and final manuscript preparation. All authors read and approved the final manuscript.

\section*{Competing Interests}
The authors declare no competing interests.

\bibliography{sample}



\section*{Supplementary material}

# Democratising Clinical AI through Dataset Condensation for Classical Clinical Models

# Supplementary Document

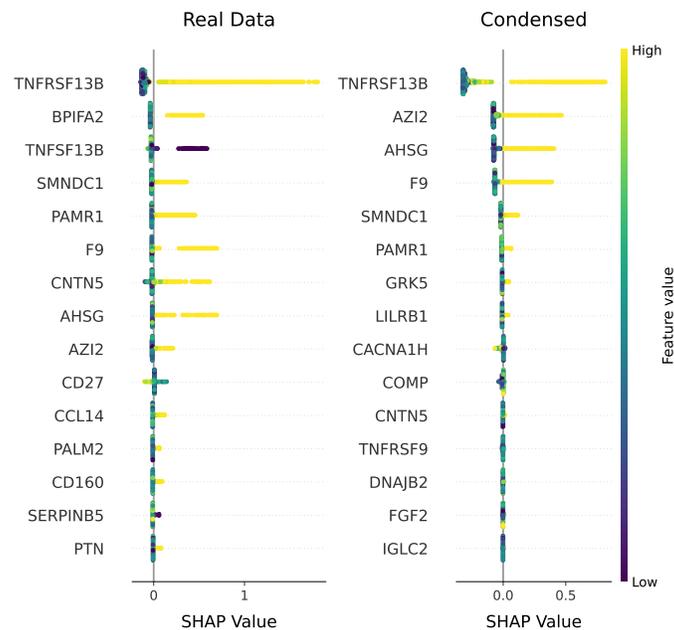

**Figure S1.** SHAP-based explainability comparison between models trained on real and condensed datasets for multiple myeloma prediction using the UK Biobank proteomics dataset. The best-performing condensed dataset was used for this analysis.

## Supplementary Note 1  Dataset Details

**CURIAL Datasets**
The CURIAL database is an anonymized database with United Kingdom National Health Service (NHS) approval via the national oversight/regulatory body, the Health Research Authority (HRA)(CURIAL; NHS HRA IRAS ID: 281832).

Data from Oxford University Hospitals (OUH) studied here are available from the Infections in Oxfordshire Research Database[1], subject to an application meeting the ethical and governance requirements of the Database. Data from University Hospital Birmingham (UHB) and Portsmouth University Hospitals (PUH) are available on reasonable request to the respective trusts, subject to HRA requirements.
Tables S1 documents the features and number of examples at each CURIAL site.

**Diabetes Dataset**
The following features were used in the experiments with the UK Biobank diabetes dataset:

---

[1] https://oxfordbrc.nihr.ac.uk/research-themes/modernising-medical-microbiology-and-big-infection-diagnostics/infections-in-oxfordshire-research-database-iord/

1

Table S1. Clinical predictors considered for COVID-19 status prediction.

| Category | Features |
|---|---|
| Vital Signs | Heart rate, respiratory rate, oxygen saturation, systolic blood pressure, diastolic blood pressure, temperature |
| Blood Tests | Haemoglobin, haematocrit, mean cell volume, white cell count, neutrophil count, lymphocyte count, monocyte count, eosinophil count, basophil count, platelets |
| Liver Function Tests & C-reactive protein | Albumin, alkaline phosphatase, alanine aminotransferase, bilirubin, C-reactive protein |
| Urea & Electrolytes | Sodium, potassium, creatinine, urea, estimated glomerular filtration rate |

- Prev. Hypertension
- Age
- BMI
- Deprivation Index
- Gamma-Glutamyl Transferase
- Low-Density Lipoprotein
- Triglycerides
- IGF-1
- Creatinine
- Sex: Female
- Sex: Male
- Ethnicity: Black
- Ethnicity: Mixed
- Ethnicity: Other
- Ethnicity: South Asian
- Ethnicity: Unanswered
- Ethnicity: White
- Smoker: Current
- Smoker: Never
- Smoker: Previous

**SEER dataset**

The following features were used from the SEER Breast Cancer dataset:

- Age at diagnosis
- Tumor size (in mm)
- Number of regional lymph nodes examined
- Number of regional lymph nodes positive
- Race: Black
- Race: Other
- Race: White
- Marital Status: Divorced
- Marital Status: Married
- Marital Status: Separated
- Marital Status: Single (never married)
- Marital Status: Widowed
- Tumor Stage (T): T1 (2 cm)
- Tumor Stage (T): T2 (>2 cm and 5 cm)
- Tumor Stage (T): T3 (>5 cm)
- Tumor Stage (T): T4 (extension into chest wall/skin)
- Node Stage (N): N1 (13 positive lymph nodes)
- Node Stage (N): N2 (49 positive lymph nodes)
- Node Stage (N): N3 (10 positive lymph nodes)
- AJCC 6th Edition Stage: IIA
- AJCC 6th Edition Stage: IIB
- AJCC 6th Edition Stage: IIIA
- AJCC 6th Edition Stage: IIIB
- AJCC 6th Edition Stage: IIIC
- Tumor Grade II (moderately differentiated)
- Tumor Grade III (poorly differentiated)
- Tumor Grade IV (undifferentiated/anaplastic)
- Tumor Grade I (well differentiated)
- SEER Summary Stage: Distant (metastatic)
- SEER Summary Stage: Regional (locally advanced)
- Estrogen Receptor: Negative
- Estrogen Receptor: Positive
- Progesterone Receptor: Negative
- Progesterone Receptor: Positive

## Proteomics Dataset

The information regarding the available proteins can be found at [https://biobank.ndph.ox.ac.uk/ukb/label.cgi?id=1839](https://biobank.ndph.ox.ac.uk/ukb/label.cgi?id=1839).



\end{document}